\documentclass{article} 
\usepackage{main,times}
\usepackage{booktabs}

\usepackage{multirow}

\usepackage{amsmath,amsfonts,bm}

\def\eqref#1{equation~\ref{#1}}

\def\1{\bm{1}}

\DeclareMathAlphabet{\mathsfit}{\encodingdefault}{\sfdefault}{m}{sl}
\SetMathAlphabet{\mathsfit}{bold}{\encodingdefault}{\sfdefault}{bx}{n}

\DeclareMathOperator*{\argmax}{arg\,max}

\usepackage{hyperref}
\usepackage{url}
\usepackage{multirow}
\usepackage{amsmath}
\usepackage{pifont}

\usepackage{caption}
\usepackage{subcaption}
\usepackage{graphicx}
\usepackage[british,UKenglish,USenglish,american]{babel}
\usepackage[ruled,vlined]{algorithm2e}
\usepackage{float}

\newcommand\Tstrut{\rule{0pt}{2.5ex}}   
\newcommand\Bstrut{\rule[-1ex]{0pt}{0pt}}
\newcommand\TBstrut{\Tstrut\Bstrut}
\newcommand{\cmark}{\text{\ding{51}}}
\newcommand{\xmark}{\text{\ding{55}}}

\title{Auto Seg-Loss: Searching Metric Surrogates \\ for Semantic Segmentation}

\author{Hao Li$^{1*\dag}$, Chenxin Tao$^{2}$\thanks{Equal contribution. $^{\dag}$This work is done when Hao Li and Chenxin Tao are interns at SenseTime Research.}~~$^{\dag}$, Xizhou Zhu$^{3}$, Xiaogang Wang$^{1,3}$, Gao Huang$^{2}$, Jifeng Dai$^{3,4}$ \\
$^{1}$The Chinese University of Hong Kong \quad $^{2}$Tsinghua University \\
$^{3}$SenseTime Research \quad $^{4}$Qing Yuan Research Institute, Shanghai Jiao Tong University
\\
\texttt{haoli@link.cuhk.edu.hk},~ \texttt{tcx20@mails.tsinghua.edu.cn} \\
\texttt{\{zhuwalter, daijifeng\}@sensetime.com}
\\
\texttt{xgwang@ee.cuhk.edu.hk},~ \texttt{gaohuang@tsinghua.edu.cn}
}

\renewcommand{\headrulewidth}{0pt}

\begin{document}

\maketitle

\begin{abstract}

Designing proper loss functions is essential in training deep networks. Especially in the field of semantic segmentation, various evaluation metrics have been proposed for diverse scenarios. Despite the success of the widely adopted cross-entropy loss and its variants, the mis-alignment between the loss functions and evaluation metrics degrades the network performance. Meanwhile, manually designing loss functions for each specific metric requires expertise and significant manpower. In this paper, we propose to automate the design of metric-specific loss functions by searching differentiable surrogate losses for each metric. We substitute the non-differentiable operations in the metrics with parameterized functions, and conduct parameter search to optimize the shape of loss surfaces. Two constraints are introduced to regularize the search space and make the search efficient. Extensive experiments on PASCAL VOC and Cityscapes demonstrate that the searched surrogate losses outperform the manually designed loss functions consistently. The searched losses can generalize well to other datasets and networks. Code shall be released.

\end{abstract}

\section{Introduction}
\label{intro}

Loss functions are of indispensable components in training deep networks, as they drive the feature learning process for various applications with specific evaluation metrics. However, most metrics, like the commonly used 0-1 classification error, are non-differentiable in their original forms and cannot be directly optimized via gradient-based methods. Empirically, the cross-entropy loss serves well as an effective surrogate objective function for a variety of tasks concerning categorization. This phenomenon is especially prevailing in image semantic segmentation, where various evaluation metrics have been designed to address the diverse task focusing on different scenarios. Some metrics measure the accuracy on the whole image, while others focus more on the segmentation boundaries. Although cross-entropy and its variants work well for many metrics, the mis-alignment between network training and evaluation still exist and inevitably leads to performance degradation.

Typically, there are two ways for designing metric-specific loss functions in semantic segmentation. The first is to modify the standard cross-entropy loss to meet the target metric~\citep{ronneberger2015u, wu2016bridging}. The other is to design other clever surrogate losses for specific evaluation metrics~\citep{rahman2016optimizing, milletari2016v}. Despite the improvements, these handcrafted losses need expertise and are non-trivial to extend to other evaluation metrics. 

In contrast to designing loss functions manually, an alternative approach is to find a framework that can design proper loss functions for different evaluation metrics in an automated manner, motivated by recent progress in AutoML~\citep{zoph2016neural,pham2018efficient,liu2018darts, li2019lfs}. Although automating the design process for loss functions is attractive, it is non-trivial to apply an AutoML framework to loss functions. Typical AutoML algorithms require a proper search space, in which some search algorithms are conducted. Previous search spaces are either unsuitable for loss design, or too general to be searched efficiently. Recently \citet{li2019lfs} and \citet{wang2020loss} proposed search spaces based on existing handcrafted loss functions. And the algorithm searches for the best combination. However, these search spaces are still limited to the variants of cross-entropy loss, and thus do not address the mis-alignment problem well.

In this paper, we propose a general framework for searching surrogate losses for mainstream non-differentiable segmentation metrics. The key idea is that we can build the search space according to the form of evaluation metrics. In this way, the training criteria and evaluation metrics are unified. Meanwhile, the search space is compact enough for efficient search. Specifically, the metrics are first relaxed to the continuous domain by substituting the one-hot prediction and logical operations, which are the non-differentiable parts in most metrics, with their differentiable approximations. Parameterized functions are introduced to approximate the logical operations, ensuring that the loss surfaces are smooth while accurate for training. The loss parameterization functions can be of arbitrary families defined on $[0, 1]$. Parameter search is further conducted on the chosen family so as to optimize the network performance on the validation set with the given evaluation metric. Two essential constraints are introduced to regularize the parameter search space. We find that the searched surrogate losses can effectively generalize to different networks and datasets. Extensive experiments on Pascal VOC~\citep{everingham2015pascal} and Cityscapes~\citep{cordts2016cityscapes} show our approach
delivers accuracy superior than the existing losses specifically designed for individual segmentation metrics with a mild computational overhead.

Our contributions can be summarized as follows: 1) Our approach is the first general framework of surrogate loss search for mainstream segmentation metrics. 2) We propose an effective parameter regularization and parameter search algorithm, which can find loss surrogates optimizing the target metric performance with mild computational overhead. 3) The surrogate losses obtained via the proposed searching framework promote our understandings on loss function design and by themselves are novel contributions, because they are different from existing loss functions specifically
designed for individual metrics, and are transferable across different datasets and networks.

\section{Related Work}
\label{related}

\textbf{Loss function design} is an active topic in deep network training~\citep{ma2020segmentation}. In the area of image semantic segmentation, cross-entropy loss is widely used~\citep{ronneberger2015u, chen2018encoder}. But the cross-entropy loss is designed for optimizing the global accuracy measure~\citep{rahman2016optimizing, patel2020learning}, which is not aligned with many other metrics. Numerous studies are conducted to design proper loss functions for the prevalent evaluation metrics. For the mIoU metric, many works~\citep{ronneberger2015u, wu2016bridging} incorporate class frequency to mitigate the class imbalance problem. For the boundary F1 score, the losses at boundary regions are up-weighted~\citep{caliva2019distance, qin2019basnet}, so as to deliver more accurate boundaries. These works carefully analyze the property of specific evaluation metrics, and design the loss functions in a fully handcrafted way, which needs expertise. By contrast, we propose a unified framework for deriving parameterized surrogate losses for various evaluation metrics. Wherein, the parameters are searched by reinforcement learning in an automatic way. The networks trained with the searched surrogate losses deliver accuracy on par or even superior than those with the best handcrafted losses.

\textbf{Direct loss optimization} for non-differentiable evaluation metrics has long been studied for structural SVM models~\citep{joachims2005support, yue2007support, ranjbar2012optimizing}. However, the gradients w.r.t. features cannot be derived from these approaches. Therefore, they cannot drive the training of deep networks through back-propagation. \citet{hazan2010direct} proposes to optimize structural SVM with gradient descent, where loss-augmented inference is applied to get the gradients of the expectation of evaluation metrics. \citet{song2016training} further extends this approach to non-linear models (e.g., deep neural networks). However, the computational complexity is very high during each step in gradient descent. Although \citet{song2016training} and \citet{mohapatra2018efficient} have designed efficient algorithms for the Average Precision (AP) metric, other metrics still need specially designed efficient algorithms. Our method, by contrast, is general for the mainstream segmentation metrics. Thanks to the good generalizability, our method only needs to perform the search process once for a specific metric, and the searched surrogate loss can be directly used henceforth. Applying the searched loss for training networks brings very little additional computational cost.

\textbf{Surrogate loss} is introduced to derive loss gradients for the non-differentiable evaluation metrics. There are usually two ways for designing surrogate losses. The first is to handcraft an approximated differentiable metric function. For the IoU measure,  \citet{rahman2016optimizing} propose to approximate the intersection and union seperately using the softmax probabilities in a differentiable form, and show its effectiveness on binary segmentation tasks. \citet{berman2018lovasz} further deal with multi-class segmentation problems by extending mIoU from binary inputs to the continuous domain with the convex Lov\`asz extension, and their method outperforms standard cross entropy loss in multi-class segmentation tasks. For the F1 measure, dice loss is proposed by \citet{milletari2016v} as a direct objective by substituting the binary prediction with the softmax probability. In spite of the success, they do not apply for other metrics. 

The second solution is to train a network to approximate the target metric. \citet{nagendar2018neuro} train a network to approximate mIoU. \citet{patel2020learning} design a neural network to learn embeddings for predictions and ground truths for tasks other than segmentation. This line of research focuses on minimizing the approximation error w.r.t. the target metrics. But there is no guarantee that their approximations provide good loss signals for training. These approximated losses are just employed in a post-tuning setup, still relying on cross-entropy pre-trained models. Our method significantly differs in that we search surrogate losses to directly optimize the evaluation metrics in applications.

\textbf{AutoML} is a long-pursued target of machine learning~\citep{he2019automl}. Recently a sub-field of AutoML, neural architecture search~(NAS), has attracted much attention due to its success in automating the process of neural network architecture design~\citep{zoph2016neural,pham2018efficient,liu2018darts}. As an essential element, loss function has also raised the interest of researchers to automate its design process. \citet{li2019lfs} and \citet{wang2020loss} design search spaces based on existing human-designed loss functions and search for the best combination parameters. There are two issues: a) the search process outputs whole network models rather than loss functions. For every new network or dataset, the expensive search procedure is conducted again, and b) the search space are filled with variants of cross-entropy, which cannot solve the mis-alignment between cross-entropy loss and many target metrics. By contrast, our method outputs the searched surrogate loss functions of close form with the target metrics, which are transferable between networks and datasets.

\section{Revisiting Evaluation Metrics for Semantic Segmentation }
\label{revisiting}

Various evaluation metrics are defined for semantic segmentation, to address the diverse task focusing on different scenarios. Most of them are of three typical classes: Acc-based, IoU-based, and F1-score-based. This section revisits the evaluation metrics, under a unified notation set.

Table \ref{table-metrics} summarizes the mainstream evaluation metrics. The notations are as follows: suppose the validation set is composed of $N$ images, labeled with categories from $C$ classes (background included). Let $I_n, n\in \{1,\ldots,N\}$ be the $n$-th image, and $Y_n$ be the corresponding ground-truth segmentation mask. Here $Y_n = \{y_{n,c,h,w}\}_{c,h,w}$ is a one-hot vector, where $y_{n,c,h,w}\in \{0,1\}$ indicates whether the pixel at spatial location $(h, w)$ belongs to the $c$-th category ($c\in \{1,\ldots,C\}$). In evaluation, the ground-truth segmentation mask $Y_n$ is compared to the network prediction $\hat{Y}_n = \{\hat{y}_{n,c,h,w}\}_{c,h,w}$, where $\hat{y}_{n,c,h,w}\in \{0,1\}$. $\hat{y}_{n,c,h,w}$ is quantized from the continuous scores produced by the network (by argmax operation).

\textbf{Acc-based metrics.} The global accuracy measure (gAcc) counts the number of pixels correctly classified. It can be written with logical operator AND as Eq.~(\ref{gacc}). The gAcc metric counts each pixel equally, so the results of the long-tailed categories have little impact on the metric number. The mean accuracy (mAcc) metric mitigates this by normalizing within each category as in Eq.~(\ref{macc}).

\textbf{IoU-based metrics.} The evaluation is on set similarity rather than pixel accuracy. The intersection-over-union (IoU) score is evaluated between the prediction and the ground-truth mask of each category. The mean IoU (mIoU) metric averages the IoU scores of all categories, as in Eq.~(\ref{miou}).

In the variants, the frequency weighted IoU (FWIoU) metric weighs each category IoU score by the category pixel number, as in Eq.~(\ref{fwiou}). The boudary IoU (BIoU)~\citep{kohli2009robust} metric only cares the segmentation quality around the boundary, so it picks the boundary pixels out in evaluation and ignores the rest pixels. It can be calculated with Eq.~(\ref{biou}), in which $\mathrm{BD} (y_n)$ denotes the boundary region in map $y_n$. $\mathrm{BD} (y_n)$ is derived by applying XOR operation on the min-pooled ground-truth mask. The stride of the $\mathrm{Min\text{-}Pooling}(\cdot)$ is 1.

\begin{table}[tb]
\caption{Revisiting mainstream metrics for semantic segmentation. The metrics with $\dag$ measure the segmentation accuracy on the whole image. The metrics with $\ast$ focus on the boundary quality.}
\vspace{-0.5em}
\label{table-metrics}
\begin{center}
\resizebox{.95\textwidth}{!}{
\begin{tabular}{l|l|l}
\hline
\multicolumn{1}{c|}{\bf Type} & \multicolumn{1}{c|}{\bf Name} & \multicolumn{1}{c}{\bf Formula} \TBstrut\\
\hline

\multirow{4}*{Acc-based}  & Global Accuracy$^\dag$ & \parbox{11.5cm}{
\begin{equation}
\label{gacc}
    \mathrm{gAcc}=\frac{\sum_{n,c,h,w}\hat{y}_{n,c,h,w}~\mathrm{AND}~y_{n,c,h,w}}{\sum_{n,c,h,w} y_{n,c,h,w}}
\end{equation}
} \rule{0pt}{5ex}\\

\cline{2-3}

~ & Mean Accuracy$^\dag$ & \parbox{11.5cm}{
\begin{equation}
\label{macc}
    \mathrm{mAcc}=\frac{1}{C} \sum_c\frac{\sum_{n,h,w}\hat{y}_{n,c,h,w}~\mathrm{AND}~y_{n,c,h,w}}{\sum_{n,h,w} y_{n,c,h,w}}
\end{equation}
} \rule{0pt}{5ex}\\

\hline

\multirow{8}*{IoU-based}  & Mean IoU$^\dag$ & \parbox{11.5cm}{
\begin{equation}
\label{miou}
    \mathrm{mIoU}=\frac{1}{C}\sum_c\frac{\sum_{n,h,w}\hat{y}_{n,c,h,w}~\mathrm{AND}~y_{n,c,h,w}}{\sum_{n,h,w}\hat{y}_{n,c,h,w}~\mathrm{OR}~y_{n,c,h,w}}
\end{equation}
} \rule{0pt}{5ex}\\

\cline{2-3}

~ & Frequency Weighted IoU$^\dag$ & \parbox{11.5cm}{
\begin{equation}
\label{fwiou}
    \mathrm{FWIoU} = \sum_c \frac{\sum_{n,h,w} y_{n,c,h,w}}{\sum_{n,c',h,w} y_{n,c',h,w}} \frac{\sum_{n,h,w}\hat{y}_{n,c,h,w}~\mathrm{AND}~ y_{n,c,h,w}}{ \sum_{n,h,w} \hat{y}_{n,c,h,w}~\mathrm{OR}~y_{n,c,h,w}}
\end{equation}
} \rule{0pt}{5ex}\\

\cline{2-3}

~ & Boundary IoU$^\ast$ & \parbox{11.5cm}{
\begin{equation}
\begin{gathered}
\label{biou}
    \mathrm{BIoU} = \frac{1}{C} \sum_c \frac{\sum_{n}\sum_{h,w \in \mathrm{BD} (y_n)}\hat{y}_{n,c,h,w}~\mathrm{AND}~y_{n,c,h,w}}{\sum_{n}\sum_{h,w \in \mathrm{BD} (y_n)}\hat{y}_{n,c,h,w}~\mathrm{OR}~y_{n,c,h,w}}\\
    \mathrm{where}~\mathrm{BD}(y) = y~\mathrm{XOR}~\mathrm{Min\text{-}Pooling}(y)
\end{gathered}
\end{equation}
} \rule{0pt}{7ex}\\

\hline

F1-score-based & Boundary F1 Score$^\ast$ & \parbox{11.5cm}{
\begin{equation}
\label{bf1-score}
\begin{gathered}
    \mathrm{BF1\text{-}score} = \frac{1}{C}\sum_c \frac{2\times\mathrm{prec}_c\times\mathrm{recall}_c}{(\mathrm{prec}_c+\mathrm{recall}_c)} \\
    \mathrm{where}~\mathrm{prec}_c = \frac{\sum_{n,h,w} \mathrm{BD}(\hat{y}_n)_{c,h,w}~\mathrm{AND}~\mathrm{Max\text{-}Pooling}(\mathrm{BD}(y_n)_{c,h,w})}{\sum_{n,h,w} \mathrm{BD}(\hat{y}_n)_{c,h,w}},\\
    \mathrm{recall}_c = \frac{\sum_{n,h,w} \mathrm{Max\text{-}Pooling}(\mathrm{BD}(\hat{y}_n)_{c,h,w})~\mathrm{AND} (\mathrm{BD}(y_n)_{c,h,w})}{\sum_{n,h,w} \mathrm{BD}(y_n)_{c,h,w}}
\end{gathered}
\end{equation}
} \rule{0pt}{5ex} \\
\hline
\end{tabular}
}
\end{center}
\vspace{-0.5em}
\end{table}

\textbf{F1-score-based metrics.}  
F1-score is a criterion that takes both precision and recall into consideration. A well-known metric of this type is boundary F1-score (BF1-score)~\citep{csurka2013good}, which is widely used for evaluating boundary segmentation accuracy. The computation of precision and recall in BF1-score is as in Eq.~(\ref{bf1-score}), where $\mathrm{BD}(\hat{y}_n)$ and $\mathrm{BD}(y_n)$ are derived from Eq.~(\ref{biou}). Max pooling with stride 1, $\mathrm{Max\text{-}Pooling}(\cdot)$, is applied on the boundary regions to allow error tolerance.

\section{Auto Seg-Loss Framework}
\label{method}

\begin{figure}[t]
\begin{center}
\includegraphics[width=0.99\textwidth]{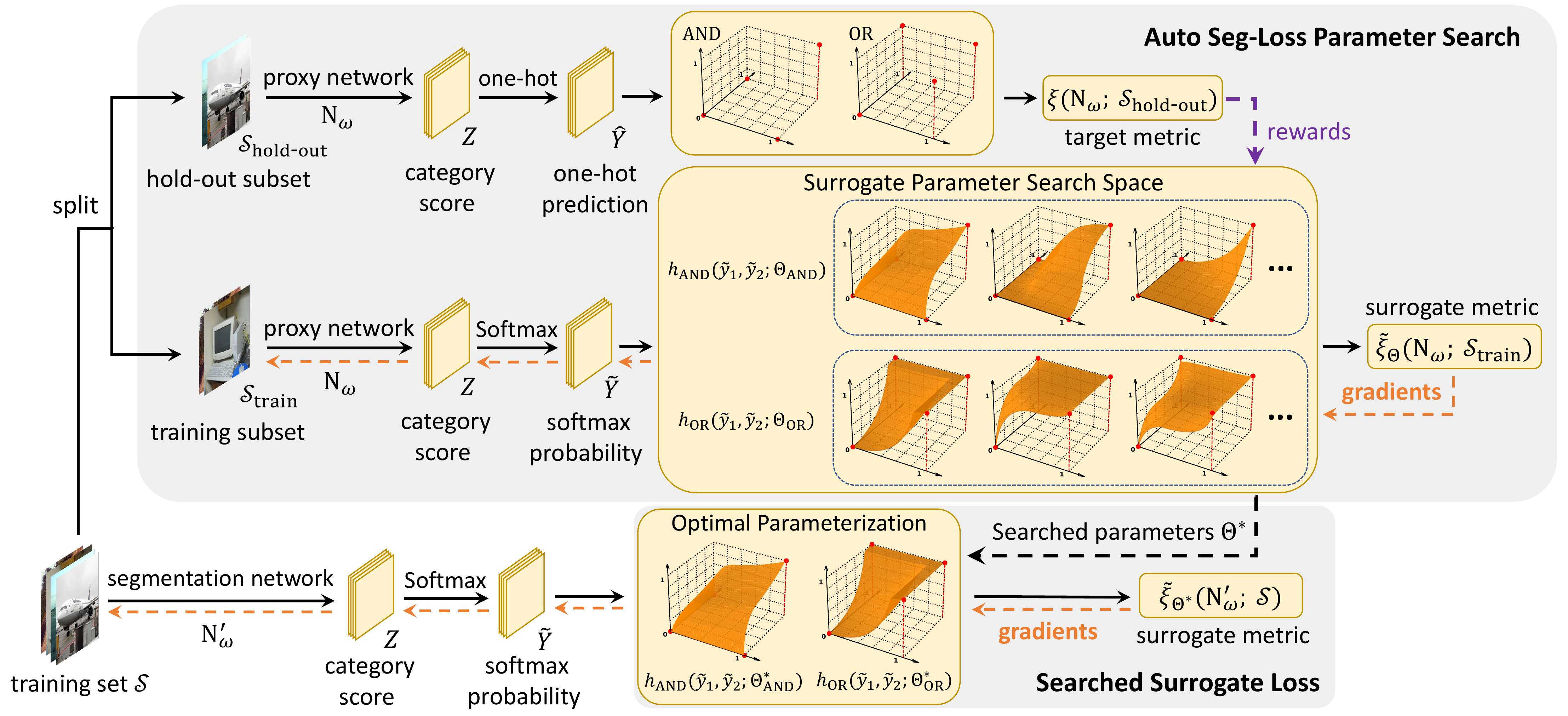}
\end{center}
\caption{Overview of the proposed Auto Seg-Loss framework. The surfaces of $h_{\mathrm{AND}}$ and $h_{\mathrm{OR}}$ shown in the "Optimal Parameterization" illustrate the searched optimal parameterization for mIoU.}
\label{fig:overview}
\end{figure}

In the Auto Seg-Loss framework, the evaluation metrics are transferred into continuous surrogate losses with learnable parameters, which are further optimized. Fig.~\ref{fig:overview} illustrates our approach.

\subsection{Extending metrics to surrogates}
\label{parameterization}
As shown in Section \ref{revisiting}, most segmentation metrics are non-differentiable because they take one-hot prediction maps as input, and contain binary logical operations. We extend these metrics to be continuous loss surrogates by smoothing the non-differentiable operations within.  

\textbf{Extending One-hot Operation.} The one-hot prediction map, $\hat{Y}_n = \{\hat{y}_{n,c,h,w}\}_{c,h,w}$, is derived by picking the highest scoring category at each pixel, which is further turned into one-hot form. Here, we approximate the one-hot predictions with softmax probabilities, as, 
\begin{equation}
\hat{y}_{n,c,h,w} \approx \widetilde{y}_{n,c,h,w} = \mathrm{Softmax}_{c}~(z_{n,c,h,w}),
\end{equation}
where $z_{n,c,h,w}\in \mathbb{R}$ is the category score output by the network (without normalization). The approximated one-hot prediction is denoted by $\widetilde{y}_{n,c,h,w}$.

\textbf{Extending Logical Operations.} As shown in Table \ref{table-metrics}, the non-differentiable logical operations, $f_{\mathrm{AND}}(y_1, y_2)$, $f_{\mathrm{OR}}(y_1, y_2)$, and $f_{\mathrm{XOR}}(y_1, y_2)$, are of indispensable components in these metrics. Because the $\mathrm{XOR}$ operation can be constructed by $\mathrm{AND}$ and $\mathrm{OR}$, $f_{\mathrm{XOR}}(y_1, y_2) = f_{\mathrm{OR}}(y_1, y_2) - f_{\mathrm{AND}}(y_1, y_2)$, we focus on extending  $f_{\mathrm{AND}}(y_1, y_2)$ and $f_{\mathrm{OR}}(y_1, y_2)$ to the continuous domain.

Following the common practice,  the logical operators are substituted with arithmetic operators
\begin{equation}
\begin{split}
\label{equ-and-or}
f_{\mathrm{AND}}(y_1, y_2) = y_1 y_2,~ f_{\mathrm{OR}}(y_1, y_2) = y_1 + y_2 - y_1 y_2,
\end{split}
\end{equation}
where $y_1, y_2 \in \{0, 1\}$. Eq.~(\ref{equ-and-or}) can be directly extended to take continuous  $y_1, y_2 \in [0, 1]$ as inputs. By such an extension, together with the approximated one-hot operation, a na\"{\i}ve version of differentiable surrogate losses can be obtained. The strength of such surrogates is that they are directly derived from the metrics, which significantly reduces the gap between training and evaluation. However, there is no guarantee that the loss surfaces formed by na\"{\i}vely extending Eq.~(\ref{equ-and-or}) provide accurate loss signals. To adjust the loss surfaces, we parameterize the $\mathrm{AND}$ and $\mathrm{OR}$ functions as 
\begin{equation} \label{equ-fg-andor}
\begin{split}
    h_{\mathrm{AND}}(y_1, y_2; \theta_{\mathrm{AND}}) &= g(y_1; \theta_{\mathrm{AND}}) ~ g(y_2; \theta_{\mathrm{AND}}),\\
    h_{\mathrm{OR}}(y_1, y_2; \theta_{\mathrm{OR}}) &= g(y_1; \theta_{\mathrm{OR}}) + g(y_2; \theta_{\mathrm{OR}}) - g(y_1; \theta_{\mathrm{OR}}) ~ g(y_2; \theta_{\mathrm{OR}}),
\end{split}
\end{equation}
where $g(y; \theta): [0,1] \rightarrow \mathbb{R}$ is a scalar function parameterized by $\theta$.

The parameterized function $g(y; \theta)$ can be from arbitrary function families defined on $[0, 1]$, e.g., piecewise linear functions and piecewise B\'ezier curves. With a chosen function family, the parameters $\theta$ control the shape of loss surfaces. We seek to search for the optimal parameters $\theta$ so as to maximize the given evaluation metric.

Meanwhile, optimal parameter search is non-trivial. With the introduced parameters, the plasticity of loss surfaces is strong. The parameterized loss surfaces may well be chaotic, or be far away from the target evaluation metric even at the binary inputs. For more effective parameter search, we regularize the loss surfaces by introducing two constraints on $g(y; \theta)$.

\textit{Truth-table constraint} is introduced to enforce the surrogate loss surfaces taking the same values as the evaluation metric score at binary inputs. This is applied by enforcing
\begin{equation}
    g(0; \theta) = 0,~ g(1; \theta) = 1.
\end{equation}
Thus, the parameterized functions $h(y_1, y_2; \theta)$ preserve the behavior of the corresponding logical operations $f(y_1, y_2)$ on binary inputs $y_1, y_2 \in \{0, 1\}$.

\textit{Monotonicity constraint} is introduced based on the observation of monotonicity tendency in the truth tables of $\mathrm{AND}$ and $\mathrm{OR}$. It pushes the loss surfaces towards a benign landscape, avoiding dramatic non-smoothness. The monotonicity constraint is enforced on $h_{\mathrm{AND}}(y_1, y_2)$ and $h_{\mathrm{OR}}(y_1, y_2)$, as
$$
{\partial h_{\mathrm{AND}}}/{\partial y_i} \geq 0,~ {\partial h_{\mathrm{OR}}}/{\partial y_i} \geq 0,~ \forall y_i \in [0, 1],~ i = 1, 2.
$$
Applying the chain rule and the truth table constraint,  the monotonicity constraint implies
\begin{equation}
    \partial g(y; \theta)/\partial y \geq 0, ~ \forall y \in [0, 1].
\end{equation}
Empirically we find it important to enforce these two constraints in parameterization.

\textbf{Extending Evaluation Metrics.} Now we can extend the metrics to surrogate losses by a) replacing the one-hot predictions with softmax probabilities, and b) substituting the logical operations with parameterized functions. Note that if the metric contains several logical operations, their parameters will not be shared. The collection of parameters in one metric are denoted as $\Theta$. For a segmentation network $\rm{N}$ and evaluation dataset $\mathcal{S}$, the score of the evaluation metric is denoted as $\xi(\rm{N}; \mathcal{S})$. And the parameterized surrogate loss is denoted as $\widetilde{\xi}_{\Theta}(\rm{N};\mathcal{S})$.

\subsection{Surrogate parameterization}
\label{search space}

The parameterized function can be from any function families defined on [0, 1], such as picewise B\'ezier curve and piecewise linear functions. Here we choose the piecewise B\'ezier curve for parameterizing $g(y; \theta)$, which is widely used in computer graphics and is easy to enforce the constraints via its control points. We also verify the effectiveness of parameterizing $g(y; \theta)$ by piecewise linear functions. See Fig.~\ref{fig-bezier-vis} for visualization and Appendix \ref{append-linear} for more details.

A piecewise B\'ezier curve consists of a series of quadratic B\'ezier curves, where the last control point of one curve segment coincides with the first control point of the next curve segment. If there are $n$ segments in a piecewise B\'ezier curve, the $k$-th segment is defined as
\begin{equation} \label{equ-bezier}
B(k,s) = (1-s)^2B_{2k} + 2s(1-s)B_{2k+1} + s^2B_{2k+2},~ 0\le s\le 1
\end{equation}
where $s$ transverses the $k$-th segment, $B_{2k+i}=(B_{(2k+i),u}, B_{(2k+i),v})~(i = 0, 1, 2)$ denotes the $i$-th control point on the $k$-th segment, in which $u, v$ index the 2-d plane axes.  A piecewise B\'ezier curve with $n$ segments has $2n+1$ control points in total. To parameterize $g(y; \theta)$, we assign
\begin{subequations}
\begin{align} 
y =&~ (1-s)^2B_{2k, u} + 2s(1-s)B_{(2k+1), u} + s^2B_{(2k+2), u}, \label{equ-para-bezier1}\\
g(y; \theta) =&~ (1-s)^2B_{2k, v} + 2s(1-s)B_{(2k+1), v} + s^2B_{(2k+2), v}, \label{equ-para-bezier2}\\
\text{s.t.} &~B_{2k, u} \le y \le B_{(2k+2), u},\label{equ-para-bezier3}
\end{align}
\end{subequations}
where $\theta$ is the control point set, $B_{2k,u} < B_{(2k+1),u} < B_{(2k+2), u}$, $ 0\le k \le n-1$. Given an input $y$, the segment index $k$ and the transversal parameter $s$ are derived from Eq.~(\ref{equ-para-bezier3}) and Eq.~(\ref{equ-para-bezier1}), respectively. Then $g(y; \theta)$ is assigned as Eq.~(\ref{equ-para-bezier2}). Because $g(y; \theta)$ is defined on $y\in [0,1]$, we arrange the control points in the $u$-axis as, $B_{0,u} = 0, ~B_{2n,u} = 1$, where the $u$-coordinate of the first and the last control points are at $0$ and $1$, respectively.

The strength of the piecewise B\'ezier curve is that the curve shape is defined explicitly via the control points. Here we enforce the truth-table and the monotonicity constraints on the control points via,
\begin{align}
    &B_{0,v} = 0, ~B_{2n,v} = 1 \tag{\emph{truth-table constraint}}; \\
    &B_{2k,v} \le B_{(2k+1),v} \le B_{(2k+2), v},\quad k=0, 1, \dots, n-1. \quad  \tag{\emph{monotonicity constraint}}
\end{align}

To fulfill the above restrictions in optimization, the specific form of the parameters is given by
$$
\theta = \left\{\left(\frac{B_{i, u} - B_{(i-1), u}}{B_{2n, u} - B_{(i-1), u}}, \frac{B_{i, v} - B_{(i-1), v}}{B_{2n, v} - B_{(i-1), v}} \right) \mid i = 1, 2, \dots, 2n-1 \right\},
$$
with $B_0=(0, 0)$ and $B_{2n}=(1, 1)$ fixed. So every $\theta_i = (\theta_{i,u}, \theta_{i,v})$ is in range $[0, 1]^2$ and it is straight-forward to compute the actual coordinates of control points from this parameterized form.
Such parameterization makes each $\theta_{i}$ independent with each other, and thus simplifies the optimization. By default, we use piecewise B\'ezier curve with two segments to parameterize $g(y, \theta)$.

\begin{minipage}{0.4\textwidth}
\begin{figure}[H]
    \begin{center}
    \vspace{-0.7cm}
    \includegraphics[width=\textwidth]{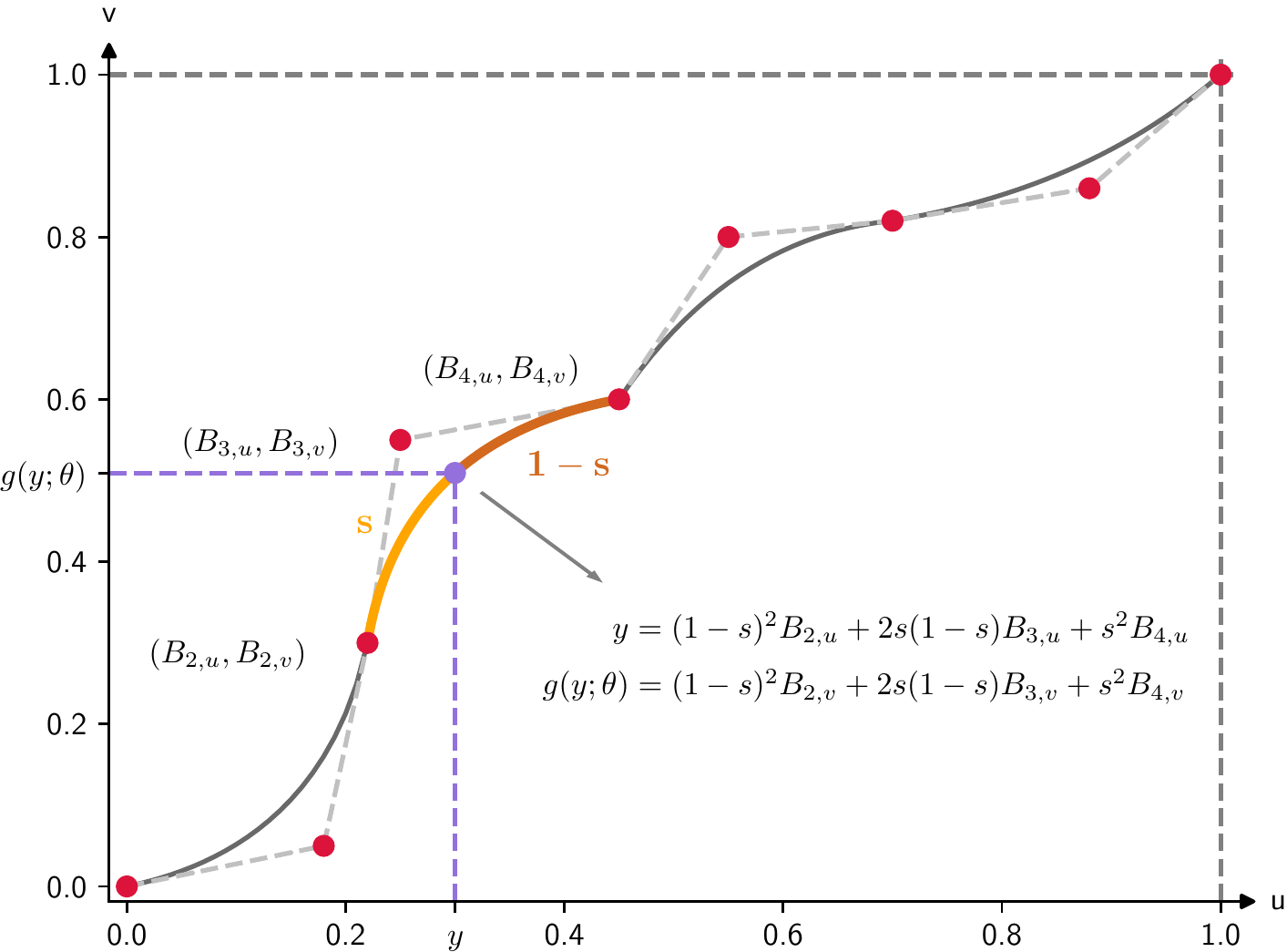}
    \end{center}
    \caption{Parameterization of $g(y; \theta)$ using Piecewise B\'ezier curve with four segments. The red points are control points. The purple point is on the curve, which shows the relationship among $y$, $g(y;\theta)$ and the transversal parameter $s$.}
    \vspace{-1em}
    \label{fig-bezier-vis}
\end{figure}
\end{minipage}
\hspace{0.5em}
\begin{minipage}{0.58\textwidth}
\begin{algorithm}[H]
\footnotesize
\SetAlgoLined
\SetKwInOut{KwInput}{Input}
\SetKw{KwReturn}{return}

\KwInput{Initialized network $\rm{N}_{\omega_0}$, initialized distribution ${\mu}_1$ and ${\sigma}^2$, target metric $\xi$, training set $\mathcal{S}_{\text{train}}$ and hold-out training set $\mathcal{S}_{\text{hold-out}}$}
  
\KwResult{Obtained optimal parameters $\Theta^*$}
\BlankLine
\For{t = 1 \KwTo T}{
    \For{i = 1 \KwTo M}{
        Sample parameter $\Theta_i^{(t)} \sim \mathcal{N}_{\text{trunc}[0, 1]}({\mu}_t,~ {\sigma}^2I)$\;
        
        Network training  $\omega^*(\Theta_i^{(t)}) = \argmax_\omega~ \widetilde{\xi}_{\Theta_i^{(t)}}(\rm{N}_\omega; \mathcal{S}_{\text{train}})$, \qquad with $w$ initialized from $w_0$\;
        
        Compute the evaluation metric score $\xi(\Theta_i^{(t)}) = \xi(\rm{N}_{\omega^*(\Theta_i^{(t)})}; \mathcal{S}_{\text{hold-out}})$\;
    }
    
    Update $\mu_{t+1} = \argmax_{\mu} \frac{1}{M}\sum_{i=1}^MR(\mu, \mu_t, \Theta_i^{(t)})$\;
}
\KwReturn{$\Theta^* = \argmax_{\mu_t} \sum_{i=1}^{M} \xi(\Theta_i^{(t)}), \forall t=1, \ldots, T+1$}

\caption{Auto Seg-Loss Parameter Search}\label{algo-1}
\end{algorithm}
\end{minipage}

\subsection{Surrogate parameter optimization}
\label{optimization}
Algorithm \ref{algo-1} describes our parameter search algorithm. The training set is split into two subsets, $\mathcal{S}_{\text{train}}$ for training and $\mathcal{S}_{\text{hold-out}}$ for evaluation in the search algorithm, respectively. Specifically, suppose we have a segmentation network $\rm{N}_\omega$ with weights $\omega$, our search target is the parameters that maximize the evaluation metric on the hold-out training set  $\xi(\rm{N}_\omega; \mathcal{S}_\text{hold-out})$
\begin{equation} \label{equ-optimization-problem}
\begin{aligned}
    \max_\Theta~ \xi(\Theta) =~ \xi(\rm{N}_{\omega^*(\Theta)}; \mathcal{S}_\text{hold-out}), \quad
    \text{s.t.} \quad \omega^*(\Theta) =~ \argmax_\omega~ \widetilde{\xi}_\Theta(\rm{N}_\omega; \mathcal{S}_{\text{train}}).
\end{aligned}
\end{equation}
To optimize Eq.~(\ref{equ-optimization-problem}), the segmentation network is trained with SGD as the inner-level problem. At the outer-level, we use reinforcement learning as our searching algorithm, following the common practice in AutoML~\citep{zoph2016neural, pham2018efficient}. Other searching algorithms, such as evolutionary algorithm, may also be employed. Specifically, the surrogate parameters are searched via the PPO2 algorithm~\citep{schulman2017proximal}. The process consists of $T$ sampling steps. In the $t$-th step, we aim to explore the search space around that from $t-1$. Here $M$ parameters $\{\Theta_i^{(t)}\}_{i=1}^{M}$ are sampled independently from a truncated normal distribution~\citep{burkardt2014truncated}, as $\Theta \sim \mathcal{N}_{\text{trunc}[0, 1]}({\mu}_t,~ {\sigma}^2I)$, with each variable in range $[0, 1]$. In it, ${\mu}_t$ and ${\sigma}^2I$ denote the mean and covariance of the parent normal distribution~($\sigma$ is fixed as 0.2 in this paper). ${\mu}_t$ summarizes the information from the $(t-1)$-th step. $M$ surrogate losses are constructed with the sampled parameters, which drive the training  of $M$ segmentation networks separately. To optimize the outer-level problem, we evaluate these models with the target metric and take the evaluation scores as rewards for PPO2. Following the PPO2 algorithm, ${\mu_{t+1}}$  is computed as $\mu_{t+1} = \argmax_{\mu} \frac{1}{M}\sum_{i=1}^{M} R(\mu, \mu_t, \Theta_i)$,where the reward $R(\mu, \mu_t, \Theta_i)$ is as
$$
R(\mu, \mu_t, \Theta_i) = \min \left(
\frac{p(\Theta_i; {\mu}, {\sigma}^2I)}{p(\Theta_i; {\mu}_{t}, {\sigma}^2I)} \xi(\Theta_i), ~
\text{CLIP}\left(\frac{p(\Theta_i; {\mu}, {\sigma}^2I)}{p(\Theta_i; {\mu}_{t}, {\sigma}^2I)} , 1 - \epsilon, 1 + \epsilon\right)\xi(\Theta_i)
\right),
$$
 where $\min(\cdot, \cdot)$ picks the smaller item from its inputs, $\text{CLIP}(x, 1-\epsilon, 1+\epsilon)$ clips $x$ to be within $1-\epsilon$ and $1+\epsilon$, and $p(\Theta_i; {\mu}, {\sigma}^2I)$ is the PDF of the truncated normal distribution. Note that the mean reward of the $M$ samples is subtracted when computing $\xi(\Theta_i)$ for better convergence. After $T$ steps,  the mean $\mu_t$ with the highest average evaluation score is output as the final parameters $\Theta^*$.

Empirically we find the searched losses have good transferability, i.e., they can be applied for different datasets and networks. Benefiting from this, we use a light proxy task for parameter search. In it, we utilize a smaller image size, a shorter learning schedule and a lightweight network. Thus, the whole search process is quite efficient~(8 hours on PASCAL VOC with 8 NVIDIA Tesla V100 GPUs). More details are in Appendix~\ref{append-implement}. In addition, the search process can be conducted only once for a specific metric and the resulting surrogate loss can be directly used for training henceforth.

\section{Experiments}
\label{exp}

We evaluate on the PASCAL VOC 2012~\citep{everingham2015pascal} and the Cityscapes~\citep{cordts2016cityscapes} datasets. We use Deeplabv3+~\citep{chen2018encoder} with ResNet-50/101~\citep{he2016deep} as the network model. During the surrogate parameter search, we randomly sample 1500 training images in PASCAL VOC and 500 training images in Cityscapes to form the hold-out set $\mathcal{S}_{\text{hold-out}}$, respectively. The remaining training images form the training set $\mathcal{S}_{\text{train}}$ in search. $\mu_0$ is set to make $g(y; \theta) = y$. The backbone network is ResNet-50. The images are down-sampled to be of $128\times 128$ resolution. SGD lasts only 1000 iterations with a mini-batch size of 32. After the search procedure, we re-train the segmentation networks with ResNet-101 using the searched losses on the full training set and evaluate them on the actual validation set. The re-train settings are the same as Deeplabv3+~\citep{chen2018encoder}, except that the loss function is substituted by the obtained surrogate loss. The search time is counted on 8 NVIDIA Tesla V100 GPUs. More details are in Appendix~\ref{append-implement}.

\subsection{Searching for Different Metrics}

In Table \ref{exp-voc&city}, we compare our searched surrogate losses against the widely-used cross-entropy loss and its variants, and some other metric-specific surrogate losses. We also seek to compare with the AutoML-based method in \citet{li2019lfs}, which was originally designed for other tasks. But we cannot get reasonable results due to convergence issues. The results show that our searched losses are on par or better the previous losses on their target metrics. It is interesting to note that the obtained surrogates for boundary metrics (such as BIoU and BF1) only focus on the boundary areas, see Appendix~\ref{append-vis} for further discussion. We also tried training segmentation networks driven by both searched mIoU and BIoU/BF1 surrogate losses. Such combined losses refine the boundaries while keeping reasonable global performance.

\setlength{\tabcolsep}{3pt}
\begin{table}[th]
\caption{Performance of different losses on PASCAL VOC and Cityscapes segmentation. The results of each loss function's target metrics are \underline{underlined}. The scores whose difference with the highest is less than 0.3 are marked in {\bf bold}.} 
\vspace{-1em}
\label{exp-voc&city}
\begin{center}
\resizebox{\textwidth}{!}{
\begin{tabular}{l|cccccc|cccccc}
\hline
\multicolumn{1}{c|}{\bf Dataset} &\multicolumn{6}{c|}{\bf PASCAL VOC} &\multicolumn{6}{c}{\bf Cityscapes}\TBstrut \\
\hline
\multicolumn{1}{c|}{\bf Loss Function} &\multicolumn{1}{c}{\bf mIoU}  &\multicolumn{1}{c}{\bf FWIoU}  &\multicolumn{1}{c}{\bf BIoU}  &\multicolumn{1}{c}{\bf BF1}  &\multicolumn{1}{c}{\bf mAcc}  &\multicolumn{1}{c|}{\bf gAcc} &\multicolumn{1}{c}{\bf mIoU}  &\multicolumn{1}{c}{\bf FWIoU}  &\multicolumn{1}{c}{\bf BIoU}  &\multicolumn{1}{c}{\bf BF1}  &\multicolumn{1}{c}{\bf mAcc}  &\multicolumn{1}{c}{\bf gAcc}\TBstrut \\
\hline
Cross Entropy          & 78.69 & 91.31 & 70.61 & 65.30 & 87.31 & \underline{95.17} & 79.97 & {\bf 93.33} & 62.07 & 62.24 & 87.01 & \underline{\bf 96.44}\Tstrut \\
WCE{\scriptsize ~\citep{ronneberger2015u}}       & 69.60 & 85.64 & 61.80 & 37.59 & \underline{\bf 92.61} & 91.11 & 73.01 & 90.51 & 53.07 & 51.19 & \underline{\bf 89.22} & 94.56 \\
DPCE{\scriptsize ~\citep{caliva2019distance}} & 79.82 & 91.76 & \underline{71.87} & \underline{66.54} & 87.76 & {\bf 95.45} & 80.27 & {\bf 93.38} & \underline{62.57} & \underline{65.99} & 86.99 & {\bf 96.46} \\
SSIM{\scriptsize ~\citep{qin2019basnet}}  & 79.26 & 91.68 & \underline{71.54} & \underline{66.35} & 87.87 & 95.38 & {\bf 80.65} & {\bf 93.22} & \underline{63.04} & \underline{72.20} & 86.88 & {\bf 96.39}\Bstrut\\
\hline
DiceLoss{\scriptsize ~\citep{milletari2016v}}     & 77.78 & 91.34 & 69.85 & 64.38 & 87.47 & 95.11 & 79.30 & {\bf 93.25} & 60.93 & 59.94 & 86.38 & {\bf 96.39}\Tstrut \\
Lov\`asz{\scriptsize ~\citep{berman2018lovasz}}  & \underline{79.72} & 91.78 & 72.47 & 66.65 & 88.64 & 95.42 & \underline{77.67} & 92.51 & 56.71 & 53.48 & 82.05 & 96.03\Bstrut \\
\hline
Searched mIoU   &\underline{\bf 80.97} &{\bf 92.09} &73.44 &68.86 &88.23 &{\bf 95.68} &\underline{\bf 80.67} &{\bf 93.30} &63.05 &67.97 &87.20 &{\bf 96.44}\Tstrut  \\
Searched FWIoU  &80.00 &\underline{\bf 91.93} &75.14 &65.67 &89.23 &95.44 &79.42 &\underline{\bf 93.33} &61.71 &59.68 &87.96 &{\bf 96.37}       \\
Searched BIoU   &48.97 &69.89 &\underline{\bf 79.27} &38.99 &81.28 &62.64 &45.89 &39.80 &\underline{\bf 63.89} &38.29 &62.80 &58.15          \\
Searched BF1   &1.93 &0.96 &7.39 &\underline{\bf 74.83} &6.51 &2.66 &6.78 &3.19 &18.37 &\underline{\bf 77.40} &12.09 &8.19         \\
Searched mAcc   &69.80 &85.86 &72.85 &35.62 &\underline{\bf 92.66} &91.28 &74.10 &90.79 &54.62 &53.45 &\underline{\bf 89.22} &94.75\\
Searched gAcc   &79.73 &91.76 & 74.09 &64.41 &88.95 &\underline{\bf 95.47} &79.41 &{\bf 93.30} &61.65 &62.04 &87.08 &\underline{\bf 96.51}\Bstrut \\
\hline
Searched mIoU + BIoU &\underline{\bf 81.19} &{\bf 92.19} &\underline{76.89} &69.56 &88.36 &{\bf 95.75} &\underline{\bf 80.43} &{\bf 93.34} &\underline{\bf 63.88} &65.87 &87.03 &{\bf 96.45}\Tstrut \\
Searched mIoU + BF1 &\underline{78.72} &90.80 &71.81 &\underline{73.57} &86.70 &94.88 &\underline{78.30} &93.00 &61.62 &\underline{71.73} &87.13 &{\bf 96.23}\Bstrut \\
\hline
\end{tabular}
}
\end{center}
\end{table}

\subsection{Generalization of the Loss}
\label{sec-generalization} 

\textbf{Generalization among datasets.}
Table \ref{exp-general-dataset} evaluates the generalization ability of our searched loss surrogates among different datasets. Due to limited computational resource, we train networks only with the searched mIoU, BF1 and mAcc surrogate losses. The results show that our searched surrogate losses generalize well between these two datasets with quite different scenes and categories.

\begin{table}[ht]
\caption{Generalization of our searched surrogate losses between PASCAL VOC and Cityscapes.}
\vspace{-1em}
\label{exp-general-dataset}
\begin{center}
\resizebox{\textwidth}{!}{
\begin{tabular}{l|cccccc|cccccc}
\hline
\multicolumn{1}{c|}{\bf Datasets} & \multicolumn{6}{c|}{\bf Cityscapes $\longrightarrow$ VOC} & \multicolumn{6}{c}{\bf VOC $\longrightarrow$ Cityscapes} \TBstrut\\
\hline
\multicolumn{1}{c|}{\bf Loss Function}  &\multicolumn{1}{c}{\bf mIoU}  &\multicolumn{1}{c}{\bf FWIoU}  &\multicolumn{1}{c}{\bf BIoU}  &\multicolumn{1}{c}{\bf BF1}  &\multicolumn{1}{c}{\bf mAcc}  &\multicolumn{1}{c|}{\bf gAcc}  &\multicolumn{1}{c}{\bf mIoU}  &\multicolumn{1}{c}{\bf FWIoU}  &\multicolumn{1}{c}{\bf BIoU}  &\multicolumn{1}{c}{\bf BF1}  &\multicolumn{1}{c}{\bf mAcc}  &\multicolumn{1}{c}{\bf gAcc} \TBstrut\\
\hline
Cross Entropy   & 78.69 & 91.31 & 70.61 & 65.30 & 87.31 & \underline{\bf 95.17} &79.97 &{\bf 93.33} &62.07 &62.24 &87.01 &\underline{{\bf 96.44}}\Tstrut\\
Searched mIoU   &\underline{\bf 80.05} &{\bf 91.72} &{\bf 73.97} &67.61 &88.01 &{\bf 95.45} &\underline{\bf 80.67} &{\bf 93.31} &{\bf 62.96} &66.48 &87.36 &{\bf 96.44} \\
Searched BF1    &1.84 &0.93 &7.42 &\underline{\bf 75.85} &6.48 &1.47 &6.67 &3.20 &19.00 &\underline{\bf 77.99} &12.12 &4.09 \\
Searched mAcc   &70.90 &86.29 &73.43 &37.18 &\underline{\bf 93.19} &91.43 &73.50 &90.68 &54.34 &54.04 &\underline{\bf 88.66} &94.68\Bstrut\\
\hline
\end{tabular}
}
\end{center}
\end{table}

\textbf{Generalization among segmentation networks.} The surrogate losses are searched with ResNet-50 + DeepLabv3+ on PASCAL VOC. The searched losses drive the training of ResNet-101 + DeepLabv3+, PSPNet~\citep{zhao2017pyramid} and HRNet~\citep{sun2019deep} on PASCAL VOC. Table \ref{exp-general-model} shows the results. The results demonstrate that our searched loss functions can be applied to various semantic segmentation networks.

\begin{table}[ht]
\caption{Generalization of our searched surrogate losses among different network architectures on PASCAL VOC. The losses are searched with ResNet-50 + DeepLabv3+ on PASCAL VOC.}
\vspace{-1em}
\label{exp-general-model}
\begin{center}
\resizebox{\textwidth}{!}{
\begin{tabular}{l|ccc|ccc|ccc|ccc}
\hline
\multicolumn{1}{c|}{\bf Network} &
\multicolumn{3}{c|}{\bf R50-DeepLabv3+} & \multicolumn{3}{c|}{\bf R101-DeepLabv3+} & \multicolumn{3}{c|}{\bf R101-PSPNet} & \multicolumn{3}{c}{\bf HRNetV2p-W48}\TBstrut\\
\hline
\multicolumn{1}{c|}{\bf Loss Function} 
&\multicolumn{1}{c}{\bf mIoU}  &\multicolumn{1}{c}{\bf BF1} &\multicolumn{1}{c|}{\bf mAcc}
&\multicolumn{1}{c}{\bf mIoU}  &\multicolumn{1}{c}{\bf BF1} &\multicolumn{1}{c|}{\bf mAcc} &\multicolumn{1}{c}{\bf mIoU}  &\multicolumn{1}{c}{\bf BF1} &\multicolumn{1}{c|}{\bf mAcc} &\multicolumn{1}{c}{\bf mIoU}  &\multicolumn{1}{c}{\bf BF1} &\multicolumn{1}{c}{\bf mAcc}\TBstrut\\
\hline
Cross Entropy   &76.22 &61.75 &85.43 & 78.69 & 65.30 & 87.31 &77.91 &64.70 &85.71 &76.35 &61.19 &85.12\Tstrut\\
Searched mIoU   &\underline{\bf 78.35} &66.93 &85.53 &\underline{\bf 80.97} &68.86 &88.23 &\underline{\bf 78.93} &65.65 &87.42 &\underline{\bf 77.26} &63.52 &86.80 \\
Searched BF1  &1.35 &\underline{\bf 70.81} &6.05 &1.43 &\underline{\bf 73.54} &6.12 &1.62 &\underline{\bf 71.84} &6.33 &1.34 &\underline{\bf 68.41} &5.99 \\
Searched mAcc  &69.82 &36.92 &\underline{\bf 91.61} &69.80 &35.62 &\underline{\bf 92.66} &71.66 &39.44 &\underline{\bf 92.06} &68.22 &35.90 &\underline{\bf 91.46}\Bstrut\\
\hline
\end{tabular}
}
\end{center}
\end{table}

\subsection{Ablation}
\label{sec-ablation}

\textbf{Parameterization and constraints.} Table \ref{exp-constraints} ablates the parameterization and the search space constraints. In it, a surrogate without parameters refers to Eq.~(\ref{equ-and-or}), with the domain extended from discrete points \{0, 1\} to continuous interval [0, 1]. This naive surrogate deliver much lower accuracy, indicating the essence of parameterization. Without the truth-table constraint, the training process diverges at the very beginning, where the loss gradients become ``NaN". And the performance drops if the monotonicity constraint is not enforced. The performance drops or even the algorithm  fails without the constraints.

\textbf{Proxy tasks for parameter search.}  Table \ref{exp-efficienty-train} ablates this. The bottom row is our default setting with a light-weight backbone, down-sampled image size and shorter learning schedule. The default setting delivers on par accuracy with heavier settings. This is consistent with the generalization ability of our surrogate losses. Thus we can improve the search efficiency via light proxy tasks.

\begin{figure}[H]
    \centering
    \begin{subfigure}{0.32\textwidth}
        \includegraphics[width=\textwidth]{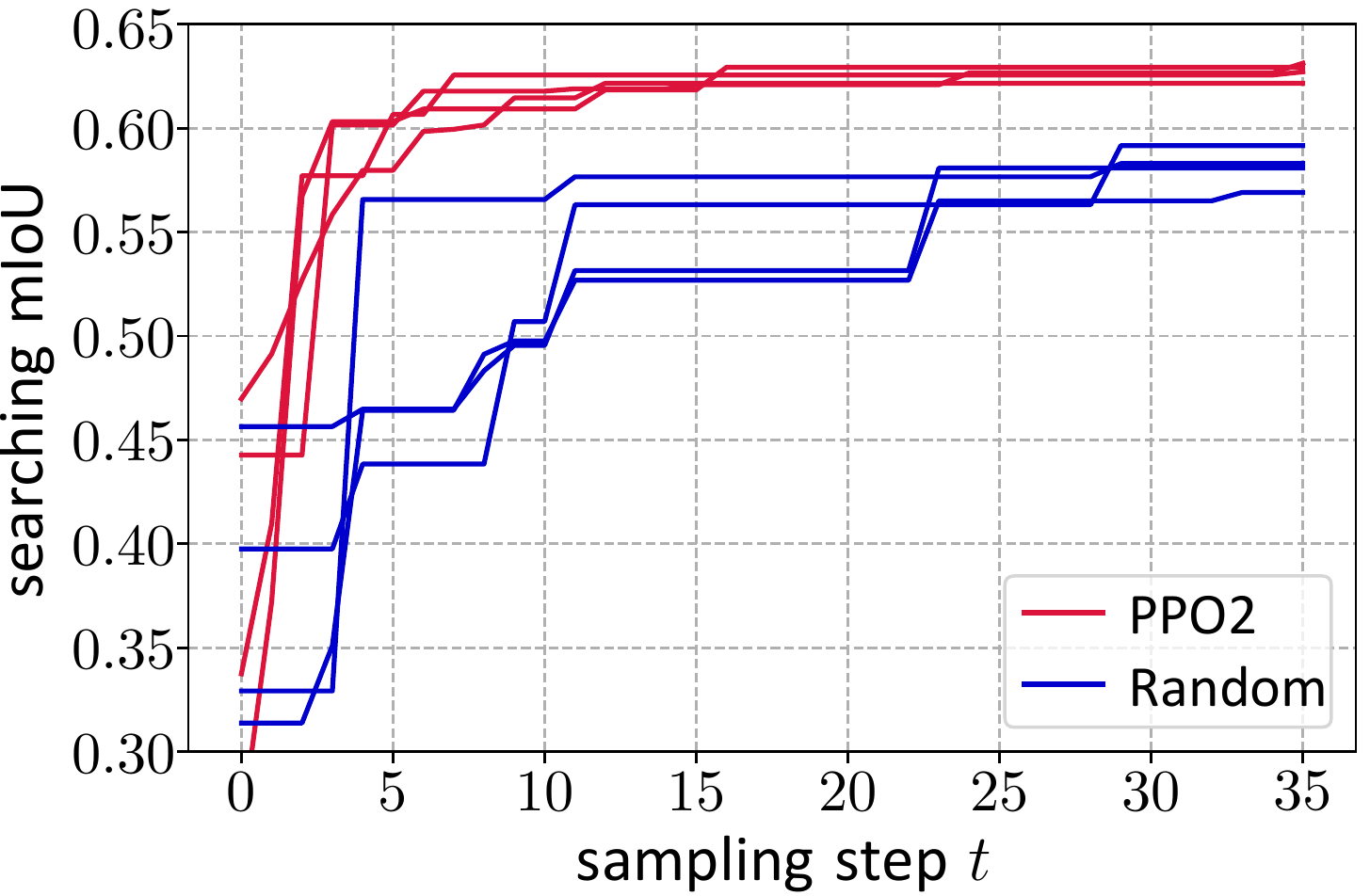}
        \caption{search for mIoU}
    \end{subfigure}
    \begin{subfigure}{0.32\textwidth}
        \includegraphics[width=\textwidth]{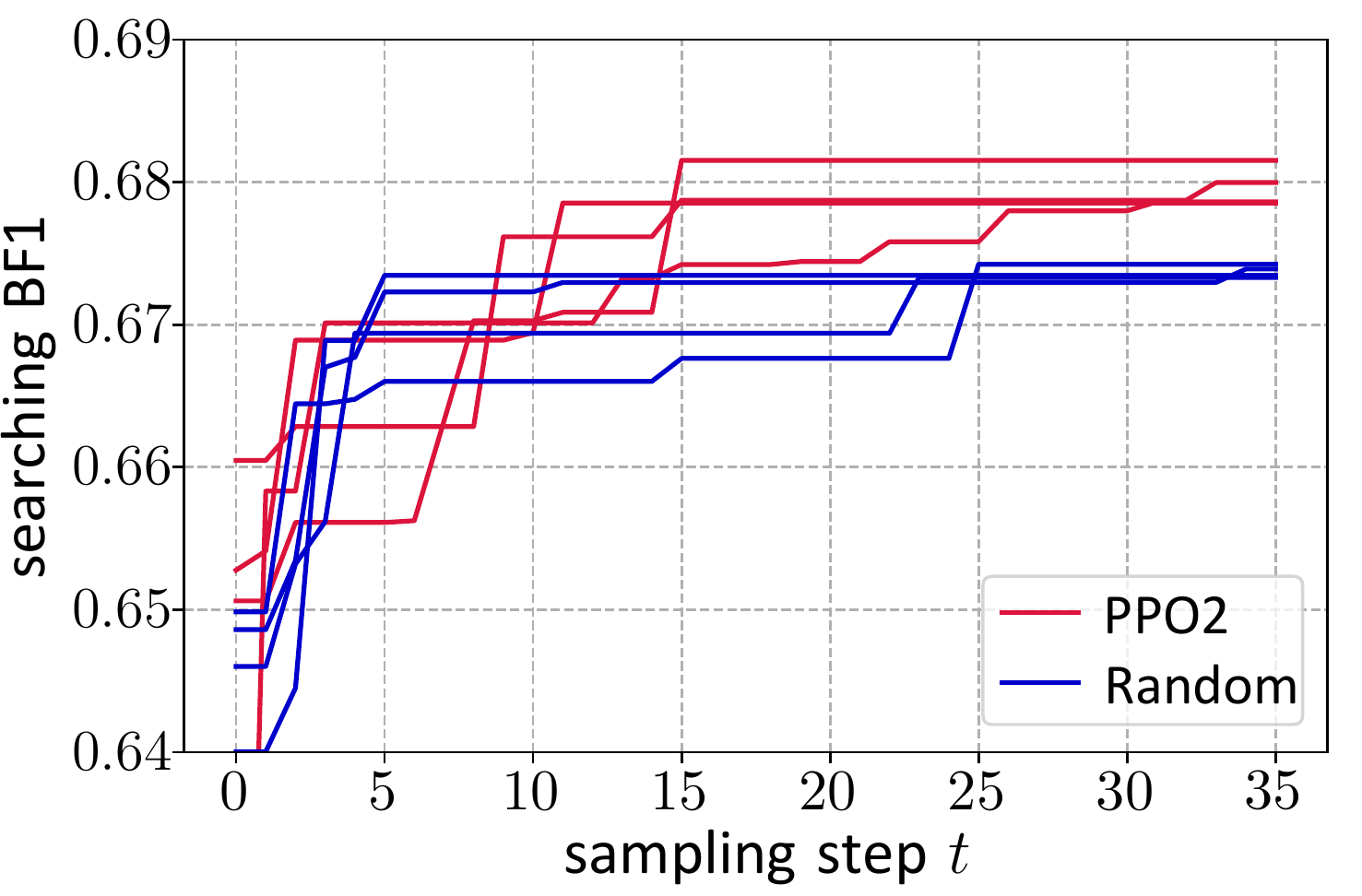}
        \caption{search for BF1}
    \end{subfigure}
    \begin{subfigure}{0.32\textwidth}
        \includegraphics[width=\textwidth]{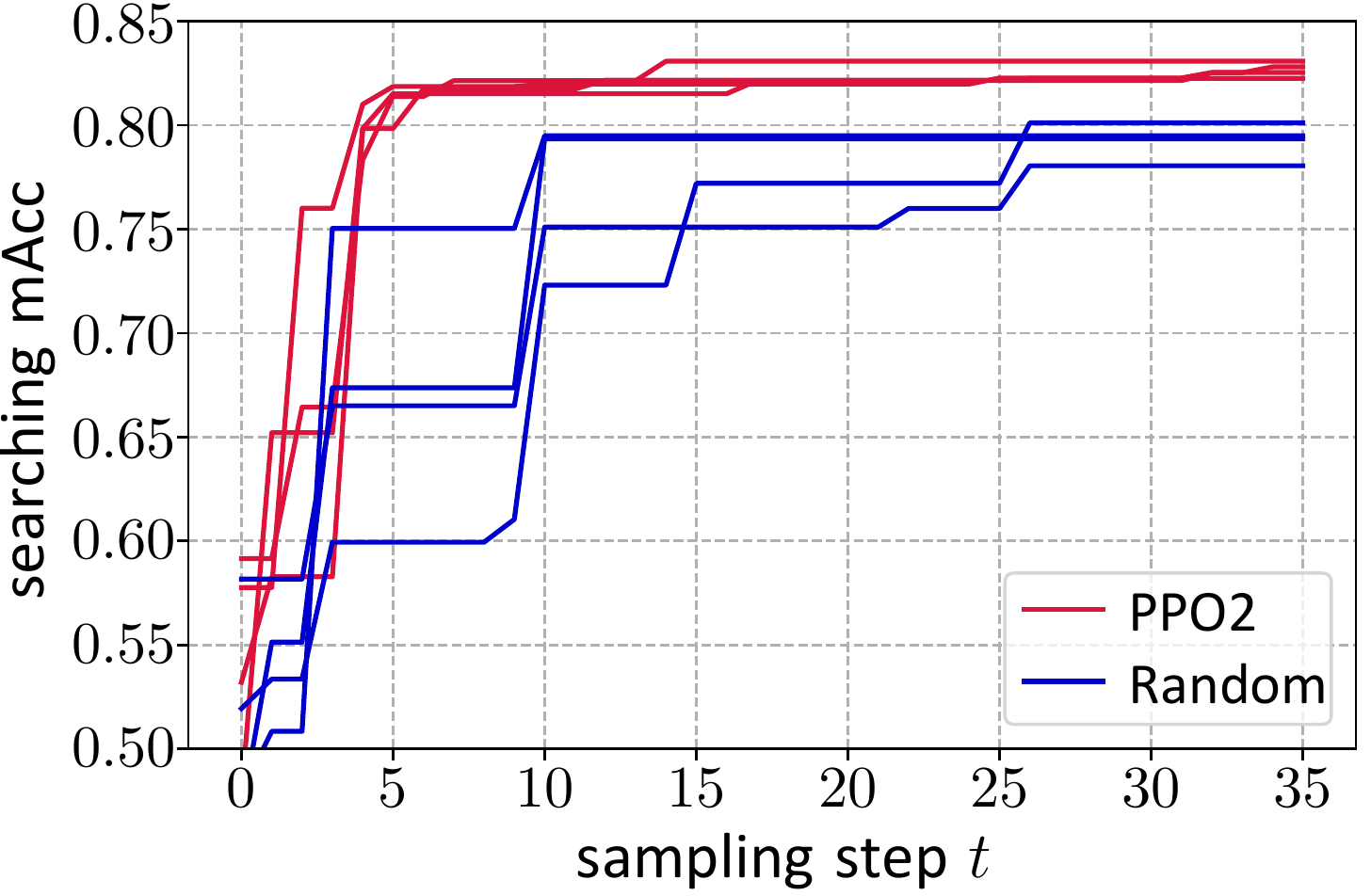}
        \caption{search for mAcc}
    \end{subfigure}
    \vspace{-0.5em}
    \caption{Ablation on loss parameter search. Each curve presents the highest average evaluation score up to the $t$-th step in one search process. The search process is repeated four times.}
    \vspace{-1.0em}
    \label{fig-search-alg}
\end{figure}

\textbf{Parameter search algorithm.} Fig.~\ref{fig-search-alg} compares the employed PPO2~\citep{schulman2017proximal} algorithm with random search. The much better performance of PPO2 suggests that surrogate loss search is non-trivial and reinforcement learning helps to improve the search efficiency.

\begin{table}[H]\centering
\begin{minipage}{0.46\textwidth}\centering
\caption{Ablation on search space constraints.}
\vspace{-0.7em}
\label{exp-constraints}
\resizebox{\textwidth}{!}{    
    \begin{tabular}{ccc|c}
    \hline
    \multicolumn{1}{c}{\bf Parameter} & \multicolumn{1}{c}{\bf Truth-table} & \multicolumn{1}{c|}{\bf Monotonicity} & \multicolumn{1}{c}{\bf VOC mIoU}\TBstrut \\
    \hline
    \xmark & \xmark & \xmark & 46.99\Tstrut\\
    \cmark & \xmark & \xmark & Fail \\
    \cmark & \cmark & \xmark & 77.76\Bstrut\\
    \hline
    \cmark & \cmark & \cmark & 80.64\TBstrut \\
    \hline
    \end{tabular}
}
\end{minipage}
\hspace{0.1cm}
\begin{minipage}{0.52\textwidth}\centering
\caption{Ablation on search proxy tasks.}
\vspace{-0.7em}
\label{exp-efficienty-train}
\resizebox{\textwidth}{!}{
    \begin{tabular}{ccc|c|c}
    \hline
    \multicolumn{1}{c}{\bf Backbone} &\multicolumn{1}{c}{\bf Image Size} &\multicolumn{1}{c}{\bf Iterations} &\multicolumn{1}{|c}{\bf Time(hours)} & \multicolumn{1}{|c}{\bf VOC mIoU} \TBstrut\\
    \hline
    R50 & 256 $\times$ 256 & 1000 & 33.0 &81.15\Tstrut \\
    R50 & 128 $\times$ 128 & 2000 & 17.1 &80.56 \\
    R101 & 128 $\times$ 128 & 1000 & 13.3 &80.75\Bstrut\\
    \hline
    R50 & 128 $\times$ 128 & 1000 & 8.5 &80.97\TBstrut\\
    \hline
    \end{tabular}
}
\end{minipage}
\end{table}

\section{Conclusion}

The introduced Auto Seg-Loss is a powerful framework to search for the parameterized surrogate losses for mainstream segmentation evalutation metrics. The non-differentiable operators are substituted by their parameterized continuous counterparts. The parameters are optimized to improve the final evaluation metrics with essential constraints. It would be interesting to extend the framework to more tasks, like object detection, pose estimation and machine translation problems.
\subsubsection*{Acknowledgments}
The work is supported by the National Key R\&D Program of China (2020AAA0105200), Beijing Academy of Artificial Intelligence and the Institute for Guo Qiang of Tsinghua University.

\clearpage
\bibliography{main}

\begin{thebibliography}{34}
\providecommand{\natexlab}[1]{#1}
\providecommand{\url}[1]{\texttt{#1}}
\expandafter\ifx\csname urlstyle\endcsname\relax
  \providecommand{\doi}[1]{doi: #1}\else
  \providecommand{\doi}{doi: \begingroup \urlstyle{rm}\Url}\fi

\bibitem[Berman et~al.(2018)Berman, Rannen~Triki, and
  Blaschko]{berman2018lovasz}
Maxim Berman, Amal Rannen~Triki, and Matthew~B Blaschko.
\newblock The lov{\'a}sz-softmax loss: A tractable surrogate for the
  optimization of the intersection-over-union measure in neural networks.
\newblock In \emph{Proceedings of IEEE Conference on Computer Vision and
  Pattern Recognition (CVPR)}, pp.\  4413--4421, 2018.

\bibitem[Burkardt(2014)]{burkardt2014truncated}
John Burkardt.
\newblock The truncated normal distribution.
\newblock \emph{Department of Scientific Computing Website, Florida State
  University}, pp.\  1--35, 2014.

\bibitem[Caliva et~al.(2019)Caliva, Iriondo, Martinez, Majumdar, and
  Pedoia]{caliva2019distance}
Francesco Caliva, Claudia Iriondo, Alejandro~Morales Martinez, Sharmila
  Majumdar, and Valentina Pedoia.
\newblock Distance map loss penalty term for semantic segmentation.
\newblock In \emph{International Conference on Medical Imaging with Deep
  Learning--Extended Abstract Track}, 2019.

\bibitem[Chen et~al.(2017)Chen, Papandreou, Kokkinos, Murphy, and
  Yuille]{chen2017deeplab}
Liang-Chieh Chen, George Papandreou, Iasonas Kokkinos, Kevin Murphy, and Alan~L
  Yuille.
\newblock Deeplab: Semantic image segmentation with deep convolutional nets,
  atrous convolution, and fully connected crfs.
\newblock \emph{IEEE Transactions on Pattern Analysis and Machine
  Intelligence}, 40\penalty0 (4):\penalty0 834--848, 2017.

\bibitem[Chen et~al.(2018)Chen, Zhu, Papandreou, Schroff, and
  Adam]{chen2018encoder}
Liang-Chieh Chen, Yukun Zhu, George Papandreou, Florian Schroff, and Hartwig
  Adam.
\newblock Encoder-decoder with atrous separable convolution for semantic image
  segmentation.
\newblock In \emph{Proceedings of the European Conference on Computer Vision
  (ECCV)}, pp.\  801--818, 2018.

\bibitem[Cordts et~al.(2016)Cordts, Omran, Ramos, Rehfeld, Enzweiler, Benenson,
  Franke, Roth, and Schiele]{cordts2016cityscapes}
Marius Cordts, Mohamed Omran, Sebastian Ramos, Timo Rehfeld, Markus Enzweiler,
  Rodrigo Benenson, Uwe Franke, Stefan Roth, and Bernt Schiele.
\newblock The cityscapes dataset for semantic urban scene understanding.
\newblock In \emph{Proceedings of IEEE Conference on Computer Vision and
  Pattern Recognition (CVPR)}, pp.\  3213--3223, 2016.

\bibitem[Csurka et~al.(2013)Csurka, Larlus, Perronnin, and
  Meylan]{csurka2013good}
Gabriela Csurka, Diane Larlus, Florent Perronnin, and France Meylan.
\newblock What is a good evaluation measure for semantic segmentation?
\newblock In \emph{Proceedings of the British Machine Vision Conference
  (BMVC)}, volume~27, pp.\  2013, 2013.

\bibitem[Everingham et~al.(2015)Everingham, Eslami, Van~Gool, Williams, Winn,
  and Zisserman]{everingham2015pascal}
Mark Everingham, SM~Ali Eslami, Luc Van~Gool, Christopher~KI Williams, John
  Winn, and Andrew Zisserman.
\newblock The pascal visual object classes challenge: A retrospective.
\newblock \emph{International Journal of Computer Vision}, 111\penalty0
  (1):\penalty0 98--136, 2015.

\bibitem[Hariharan et~al.(2011)Hariharan, Arbel{\'a}ez, Bourdev, Maji, and
  Malik]{hariharan2011semantic}
Bharath Hariharan, Pablo Arbel{\'a}ez, Lubomir Bourdev, Subhransu Maji, and
  Jitendra Malik.
\newblock Semantic contours from inverse detectors.
\newblock In \emph{Proceedings of IEEE International Conference on Computer
  Vision (ICCV)}, pp.\  991--998, 2011.

\bibitem[Hazan et~al.(2010)Hazan, Keshet, and McAllester]{hazan2010direct}
Tamir Hazan, Joseph Keshet, and David~A McAllester.
\newblock Direct loss minimization for structured prediction.
\newblock In \emph{Advances in Neural Information Processing Systems (NIPS)},
  pp.\  1594--1602, 2010.

\bibitem[He et~al.(2016)He, Zhang, Ren, and Sun]{he2016deep}
Kaiming He, Xiangyu Zhang, Shaoqing Ren, and Jian Sun.
\newblock Deep residual learning for image recognition.
\newblock In \emph{Proceedings of IEEE Conference on Computer Vision and
  Pattern Recognition (CVPR)}, pp.\  770--778, 2016.

\bibitem[He et~al.(2019)He, Zhao, and Chu]{he2019automl}
Xin He, Kaiyong Zhao, and Xiaowen Chu.
\newblock Automl: A survey of the state-of-the-art.
\newblock \emph{arXiv preprint arXiv:1908.00709}, 2019.

\bibitem[Joachims(2005)]{joachims2005support}
Thorsten Joachims.
\newblock A support vector method for multivariate performance measures.
\newblock In \emph{Proceedings of the 22nd International Conference on Machine
  Learning (ICML)}, pp.\  377--384. PMLR, 2005.

\bibitem[Kohli et~al.(2009)Kohli, Torr, et~al.]{kohli2009robust}
Pushmeet Kohli, Philip~HS Torr, et~al.
\newblock Robust higher order potentials for enforcing label consistency.
\newblock \emph{International Journal of Computer Vision}, 82\penalty0
  (3):\penalty0 302--324, 2009.

\bibitem[Li et~al.(2019)Li, Yuan, Lin, Guo, Wu, Yan, and Ouyang]{li2019lfs}
Chuming Li, Xin Yuan, Chen Lin, Minghao Guo, Wei Wu, Junjie Yan, and Wanli
  Ouyang.
\newblock Am-lfs: Automl for loss function search.
\newblock In \emph{Proceedings of the IEEE International Conference on Computer
  Vision (CVPR)}, pp.\  8410--8419, 2019.

\bibitem[Liu et~al.(2018)Liu, Simonyan, and Yang]{liu2018darts}
Hanxiao Liu, Karen Simonyan, and Yiming Yang.
\newblock Darts: Differentiable architecture search.
\newblock In \emph{Proceedings of the 6th International Conference on Learning
  Representations (ICLR)}, 2018.

\bibitem[Ma(2020)]{ma2020segmentation}
Jun Ma.
\newblock Segmentation loss odyssey.
\newblock \emph{arXiv preprint arXiv:2005.13449}, 2020.

\bibitem[Milletari et~al.(2016)Milletari, Navab, and Ahmadi]{milletari2016v}
Fausto Milletari, Nassir Navab, and Seyed-Ahmad Ahmadi.
\newblock V-net: Fully convolutional neural networks for volumetric medical
  image segmentation.
\newblock In \emph{2016 Fourth International Conference on 3D Vision (3DV)},
  pp.\  565--571. IEEE, 2016.

\bibitem[Mohapatra et~al.(2018)Mohapatra, Rolinek, Jawahar, Kolmogorov, and
  Pawan~Kumar]{mohapatra2018efficient}
Pritish Mohapatra, Michal Rolinek, CV~Jawahar, Vladimir Kolmogorov, and
  M~Pawan~Kumar.
\newblock Efficient optimization for rank-based loss functions.
\newblock In \emph{Proceedings of the IEEE Conference on Computer Vision and
  Pattern Recognition (CVPR)}, pp.\  3693--3701, 2018.

\bibitem[Nagendar et~al.(2018)Nagendar, Singh, Balasubramanian, and
  Jawahar]{nagendar2018neuro}
Gattigorla Nagendar, Digvijay Singh, Vineeth~N Balasubramanian, and CV~Jawahar.
\newblock Neuro-iou: Learning a surrogate loss for semantic segmentation.
\newblock In \emph{Proceedings of the British Machine Vision Conference
  (BMVC)}, pp.\  278, 2018.

\bibitem[Patel et~al.(2020)Patel, Hodan, and Matas]{patel2020learning}
Yash Patel, Tomas Hodan, and Jiri Matas.
\newblock Learning surrogates via deep embedding.
\newblock In \emph{Proceedings of the European Conference on Computer Vision
  (ECCV)}, 2020.

\bibitem[Pham et~al.(2018)Pham, Guan, Zoph, Le, and Dean]{pham2018efficient}
Hieu Pham, Melody Guan, Barret Zoph, Quoc Le, and Jeff Dean.
\newblock Efficient neural architecture search via parameters sharing.
\newblock In \emph{Proceedings of the 35th International Conference on Machine
  Learning (ICML)}, pp.\  4095--4104. PMLR, 2018.

\bibitem[Qin et~al.(2019)Qin, Zhang, Huang, Gao, Dehghan, and
  Jagersand]{qin2019basnet}
Xuebin Qin, Zichen Zhang, Chenyang Huang, Chao Gao, Masood Dehghan, and Martin
  Jagersand.
\newblock Basnet: Boundary-aware salient object detection.
\newblock In \emph{Proceedings of IEEE Conference on Computer Vision and
  Pattern Recognition (CVPR)}, pp.\  7479--7489, 2019.

\bibitem[Rahman \& Wang(2016)Rahman and Wang]{rahman2016optimizing}
Md~Atiqur Rahman and Yang Wang.
\newblock Optimizing intersection-over-union in deep neural networks for image
  segmentation.
\newblock In \emph{International Symposium on Visual Computing}, pp.\
  234--244. Springer, 2016.

\bibitem[Ranjbar et~al.(2012)Ranjbar, Lan, Wang, Robinovitch, Li, and
  Mori]{ranjbar2012optimizing}
Mani Ranjbar, Tian Lan, Yang Wang, Steven~N Robinovitch, Ze-Nian Li, and Greg
  Mori.
\newblock Optimizing nondecomposable loss functions in structured prediction.
\newblock \emph{IEEE Transactions on Pattern Analysis and Machine
  Intelligence}, 35\penalty0 (4):\penalty0 911--924, 2012.

\bibitem[Ronneberger et~al.(2015)Ronneberger, Fischer, and
  Brox]{ronneberger2015u}
Olaf Ronneberger, Philipp Fischer, and Thomas Brox.
\newblock U-net: Convolutional networks for biomedical image segmentation.
\newblock In \emph{International Conference on Medical Image Computing and
  Computer-Assisted Intervention}, pp.\  234--241. Springer, 2015.

\bibitem[Schulman et~al.(2017)Schulman, Wolski, Dhariwal, Radford, and
  Klimov]{schulman2017proximal}
John Schulman, Filip Wolski, Prafulla Dhariwal, Alec Radford, and Oleg Klimov.
\newblock Proximal policy optimization algorithms.
\newblock \emph{arXiv preprint arXiv:1707.06347}, 2017.

\bibitem[Song et~al.(2016)Song, Schwing, Urtasun, et~al.]{song2016training}
Yang Song, Alexander Schwing, Raquel Urtasun, et~al.
\newblock Training deep neural networks via direct loss minimization.
\newblock In \emph{Proceedings of the 33rd International Conference on Machine
  Learning (ICML)}, pp.\  2169--2177. PMLR, 2016.

\bibitem[Sun et~al.(2019)Sun, Xiao, Liu, and Wang]{sun2019deep}
Ke~Sun, Bin Xiao, Dong Liu, and Jingdong Wang.
\newblock Deep high-resolution representation learning for human pose
  estimation.
\newblock In \emph{Proceedings of IEEE Conference on Computer Vision and
  Pattern Recognition (CVPR)}, pp.\  5693--5703, 2019.

\bibitem[Wang et~al.(2020)Wang, Wang, Chi, Zhang, and Mei]{wang2020loss}
Xiaobo Wang, Shuo Wang, Cheng Chi, Shifeng Zhang, and Tao Mei.
\newblock Loss function search for face recognition.
\newblock In \emph{Proceedings of the 37th International Conference on Machine
  Learning (ICML)}. PMLR, 2020.

\bibitem[Wu et~al.(2016)Wu, Shen, and van~den Hengel]{wu2016bridging}
Zifeng Wu, Chunhua Shen, and Anton van~den Hengel.
\newblock Bridging category-level and instance-level semantic image
  segmentation.
\newblock \emph{arXiv preprint arXiv:1605.06885}, 2016.

\bibitem[Yue et~al.(2007)Yue, Finley, Radlinski, and Joachims]{yue2007support}
Yisong Yue, Thomas Finley, Filip Radlinski, and Thorsten Joachims.
\newblock A support vector method for optimizing average precision.
\newblock In \emph{Proceedings of the 30th Annual International ACM SIGIR
  Conference on Research and Development in Information Retrieval}, pp.\
  271--278, 2007.

\bibitem[Zhao et~al.(2017)Zhao, Shi, Qi, Wang, and Jia]{zhao2017pyramid}
Hengshuang Zhao, Jianping Shi, Xiaojuan Qi, Xiaogang Wang, and Jiaya Jia.
\newblock Pyramid scene parsing network.
\newblock In \emph{Proceedings of IEEE Conference on Computer Vision and
  Pattern Recognition (CVPR)}, pp.\  2881--2890, 2017.

\bibitem[Zoph \& Le(2017)Zoph and Le]{zoph2016neural}
Barret Zoph and Quoc~V. Le.
\newblock Neural architecture search with reinforcement learning.
\newblock In \emph{Proceedings of the 5th International Conference on Learning
  Representations (ICLR)}, 2017.

\end{thebibliography}
\bibliographystyle{main}

\clearpage
\appendix
\section{Implementation details}
\label{append-implement}

\textbf{Datasets.} We evaluate our approach on the PASCAL VOC 2012~\citep{everingham2015pascal} and the Cityscapes~\citep{cordts2016cityscapes} datasets. For PASCAL VOC, we follow \citet{chen2017deeplab} to augment with the extra annotations provided by \citet{hariharan2011semantic}. For Cityscapes, we follow the standard evaluation protocol in \citet{cordts2016cityscapes}.

During the surrogate parameter search, we randomly sample 1500 training images in PASCAL VOC and 500 training images in Cityscapes to form the hold-out set $\mathcal{S}_{\text{hold-out}}$, respectively. The remaining training images form the training set $\mathcal{S}_{\text{train}}$ in search.
After the search procedure, we re-train the segmentation networks with the searched losses on the full training set and evaluate them on the actual validation set.

\textbf{Implementation Details.} We use Deeplabv3+~\citep{chen2018encoder} with ImageNet-pretrained ResNet-50/101~\citep{he2016deep} backbone as the network model. The segmentation head is randomly initialized. In surrogate parameter search, the backbone is of ResNet-50. $\mu_0$ is set to make $g(y; \theta) = y$. The training and validation images are down-sampled to be of $128\times 128$ resolution. In SGD training, the mini-batch size is of 32 images, and the training is of 1000 iterations. The initial learning rate is 0.02, which is decayed by polynomial with power 0.9 and minimum learning rate 1e-4. The momentum and weight decay are set to 0.9 and 5e-4, respectively. For faster convergence, learning rate of the segmentation head is multiplied by 10. The search procedure is conducted for $T=20$ steps, and $M=32$ loss parameters are sampled in each step. In PPO2~\citep{schulman2017proximal}, the clipping threshold $\epsilon=0.1$, and $\mu_{t+1}$ is updated by 100 steps. After surrogate parameter search, the re-training settings are the same as Deeplabv3+~\citep{chen2018encoder}, except that the loss function is substituted by the searched surrogate loss function. The backbone is of ResNet-101 by default.

\section{Parameterization with piecewise linear functions}
\label{append-linear}

Here we choose the continuous piecewise linear function for parameterizing $g(y; \theta)$, where the form of constraints and parameters are very similar to that of the piecewise B\'ezier curve described in Section~\ref{search space}. Experimental results on PASCAL VOC 2012~\citep{everingham2015pascal} are presented at the end of this section. 

A continuous piecewise linear function consists of multiple line segments, where the right endpoint of one line segment coincides with the left endpoint of the next. Suppose there are $n$ line segments in a piecewise linear function, then the $k$-th line segment is defined as
\begin{equation}
A(k, s) = (1-s)A_k + sA_{k+1}, \quad 0\le s \le 1
\end{equation}
where $s$ transverses the $k$-th line segment, $A_k = (A_{k,u}, A_{k,v})$ and $A_{k+1} = (A_{(k+1),u}, A_{(k+1),v})$ are the left endpoint and right endpoint of the $k$-th line segment, respectively, in which $u, v$ index the 2-d plane axes.

To parameterize $g(y; \theta)$ via continuous piecewise linear functions, we assign
\begin{subequations}
\begin{align} 
y =&~ (1-s)A_{k,u} + sA_{(k+1),u}, \label{equ-para-linear1}\\
g(y; \theta) =&~ (1-s)A_{k,v} + sA_{(k+1),v}, \label{equ-para-linear2}\\
\text{s.t.} &~A_{k, u} \le y \le A_{(k+1), u},\label{equ-para-linear3}
\end{align}
\end{subequations}
where $\theta$ is the collection of all control points. Given an input $y$, the segment index $k$ and the transversal parameter $s$ are derived from Eq.~(\ref{equ-para-linear3}) and Eq.~(\ref{equ-para-linear1}), respectively. Then $g(y; \theta)$ is assigned as Eq.~(\ref{equ-para-linear2}).

Because $g(y; \theta)$ is a function defined on $y\in [0,1]$,  we arrange the endpoints in the $u$-axis as,
\begin{equation} \label{equ-linear-func}
A_{0,u} = 0, ~A_{n,u} = 1, ~A_{k,u} < A_{(k+1), u},~ 0\le k \le n-1
\end{equation}
where the $u$-coordinate of the first and the last endpoints are at $0$ and $1$, respectively.

We enforce the two constraints introduced in Section \ref{parameterization} on the searching space through parameters $\theta$. These two constraints can be formulated as
\begin{align}
    &A_{0, v} = 0, ~A_{n, v}=1; \tag{\emph{truth-table constraint}}\\
    &A_{k, v} \le A_{k+1, v}, ~k=0, \dots, n-1. \tag{\emph{monotonicity constraint}}
\end{align}

In practice, we divide the domain $[0, 1]$ into $n$ subintervals uniformly, and fix the $u$-coordinate of endpoints at the intersections of these intervals, i.e., $A_{k,u} = \frac{k}{n}$ where $0\le k \le n$. Then the specific form of the parameters is given by
$$
\theta = \left\{n (A_{k+1, v} - A_{k, v}) \mid k = 0, \dots, n-1 \right\}.
$$
According to the constraints, the parameters need to satisfy
$$
\frac{1}{n}\sum_{k=0}^{n-1} \theta_k = 1, ~\theta_k \ge 0, k=0, \dots, n-1.
$$
In order to meet the above constraints, during the surrogate parameter search, we first sample parameters from a normal distribution without truncation, and then apply Softmax operation on the sampled parameters. The normalized parameters are used as the actual parameters for piecewise linear functions. 

In our implementation, we use piecewise linear functions with five line segments. The effectiveness are presented in Table \ref{exp-linear-voc&city}. The searched losses parameterized with piecewise linear functions are on par or better the previous losses on their target metrics, and achieve very similar performance with that of using piecewise B\'ezier curve for the parameterization.

\setlength{\tabcolsep}{3pt}
\begin{table}[th]
\caption{Performance of different metrics on PASCAL VOC. Piecewise linear functions are used for the parameterization in our method. The results of each loss function's target metrics are \underline{underlined}. The scores whose difference with the highest is less than 0.3 are marked in {\bf bold}.}
\label{exp-linear-voc&city}
\small
\begin{center}
\begin{tabular}{l|cccccc}
\hline
\multicolumn{1}{c|}{\bf Loss Function} &\multicolumn{1}{c}{\bf mIoU}  &\multicolumn{1}{c}{\bf FWIoU}  &\multicolumn{1}{c}{\bf BIoU}  &\multicolumn{1}{c}{\bf BF1}  &\multicolumn{1}{c}{\bf mAcc}  &\multicolumn{1}{c}{\bf gAcc}\TBstrut \\
\hline
Cross Entropy          & 78.69 & 91.31 & 70.61 & 65.30 & 87.31 & \underline{{\bf 95.17}}\Tstrut \\
WCE{\scriptsize ~\citep{ronneberger2015u}}       & 69.60 & 85.64 & 61.80 & 37.59 & \underline{\bf 92.61} & 91.11 \\
DPCE{\scriptsize ~\citep{caliva2019distance}} & 79.82 & {\bf 91.76} & \underline{71.87} & \underline{66.54} & 87.76 & {\bf 95.45} \\
SSIM{\scriptsize ~\citep{qin2019basnet}}  & 79.26 & 91.68 & \underline{71.54} & \underline{66.35} & 87.87 & {\bf 95.38}\Bstrut\\
\hline
DiceLoss{\scriptsize ~\citep{milletari2016v}}     & 77.78 & 91.34 & 69.85 & 64.38 & 87.47 & 95.11\Tstrut \\
Lov\`asz{\scriptsize ~\citep{berman2018lovasz}}  & \underline{79.72} & {\bf 91.78} & 72.47 & 66.65 & 88.64 & {\bf 95.42}\Bstrut \\
\hline
Searched mIoU   & \underline{\bf 80.94} & {\bf 92.01} & 73.22 & 67.32 & 90.12 & {\bf 95.46}\Tstrut  \\
Searched FWIoU  & 79.05 & \underline{\bf 91.78} & 71.47 & 64.24 & 89.77 & {\bf 95.31}      \\
Searched BIoU   & 43.62 & 70.50 & \underline{\bf 75.37} & 46.23 & 53.21 & 82.60     \\
Searched BF1   & 1.87 & 1.03 & 6.85 & \underline{\bf 76.02} & 6.54 & 2.17      \\
Searched mAcc   & 74.33 & 88.77 & 65.96 & 46.81 & \underline{\bf 92.34} & 93.26        \\
Searched gAcc   & 78.95 & 91.51 & 69.90 & 62.65 & 88.76 & \underline{{\bf 95.19}}\Bstrut \\
\hline
Searched mIoU + BIoU & \underline{\bf 81.24} & {\bf 92.48} & \underline{\bf 75.74} & 68.19 & 90.03 & {\bf 95.42}\Tstrut \\
Searched mIoU + BF1 & \underline{79.11} & 91.38 & 71.71 & \underline{73.55} & 89.28 & {\bf 95.17}\Bstrut \\
\hline
\end{tabular}
\end{center}
\end{table}
\section{Visualization and discussion on boundary segmentation}
\label{append-vis}

During the re-training stage, we find the segmentation result trained with surrogates for BIoU and BF1 metrics particularly interesting. To further investigate their properties, we visualize the segmentation results trained with surrogates for boundary metrics.

\begin{figure}[tb]
    \centering
    \begin{subfigure}{0.45\textwidth}
        \includegraphics[width=\textwidth]{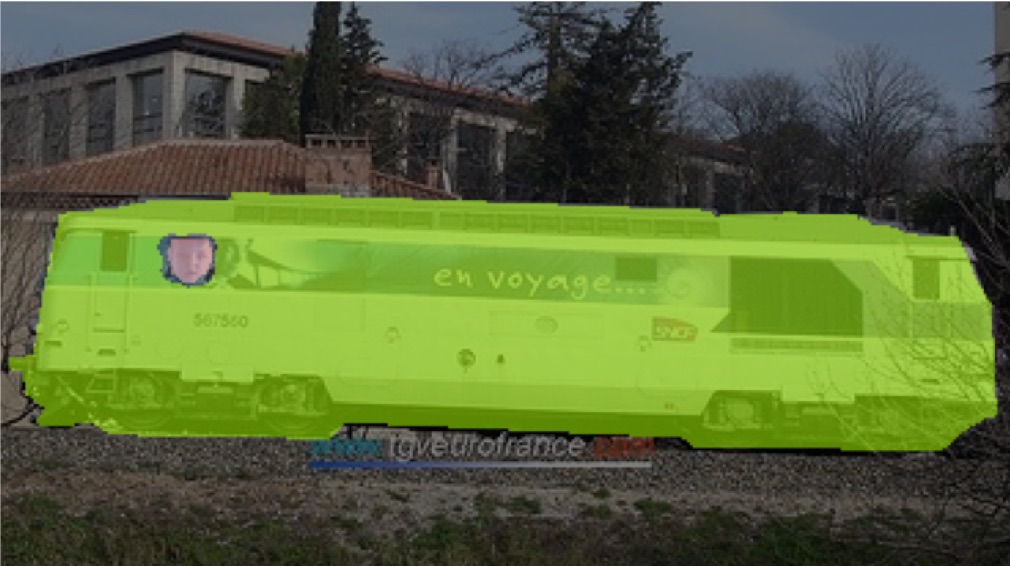}
        \caption{Ground-truth}
    \end{subfigure}
    \begin{subfigure}{0.45\textwidth}
        \includegraphics[width=\textwidth]{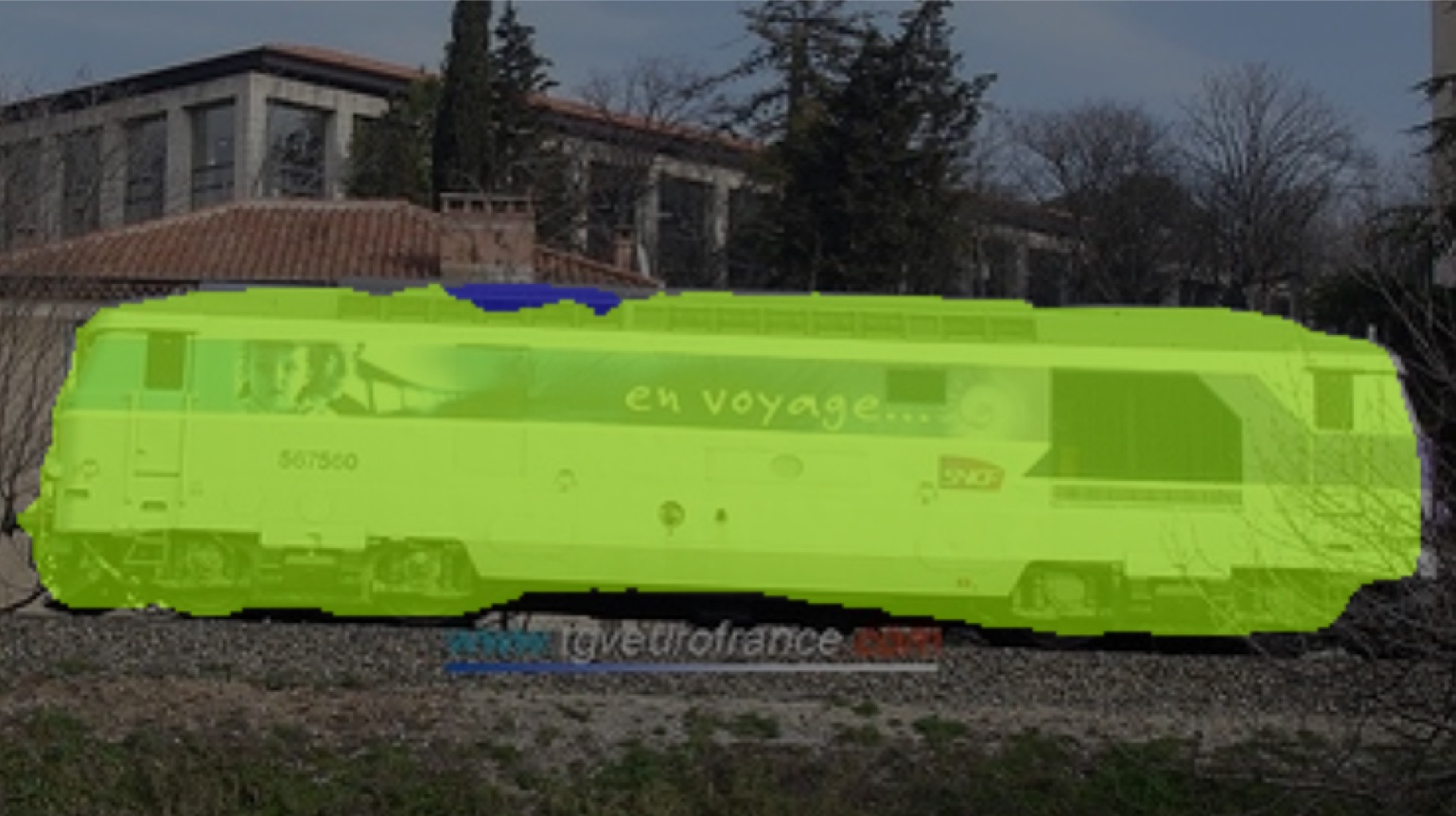}
        \caption{Searched mIoU}
    \end{subfigure}
    \begin{subfigure}{0.45\textwidth}
        \includegraphics[width=\textwidth]{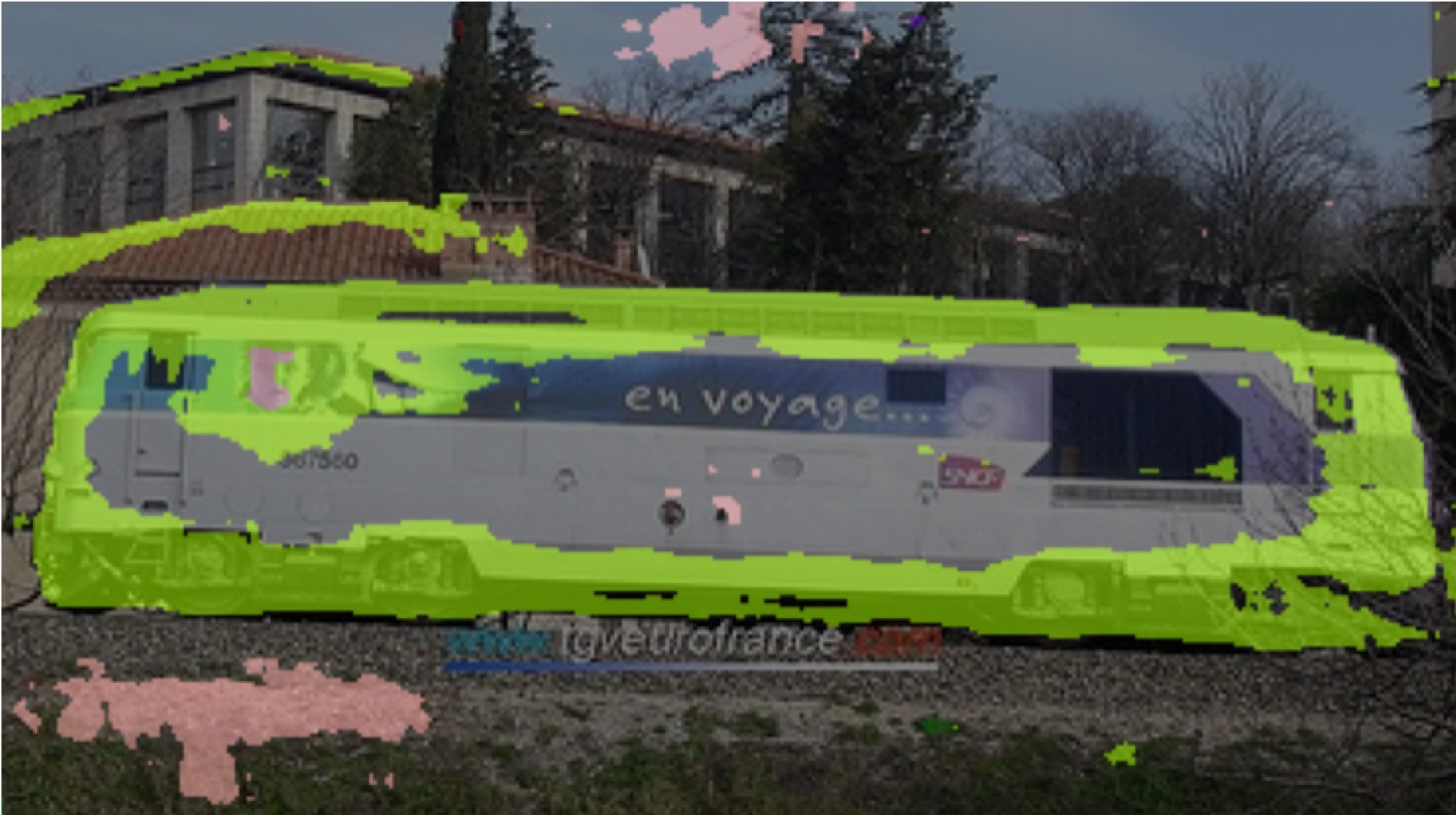}
        \caption{Searched BIoU}
    \end{subfigure}
    \begin{subfigure}{0.45\textwidth}
        \includegraphics[width=\textwidth]{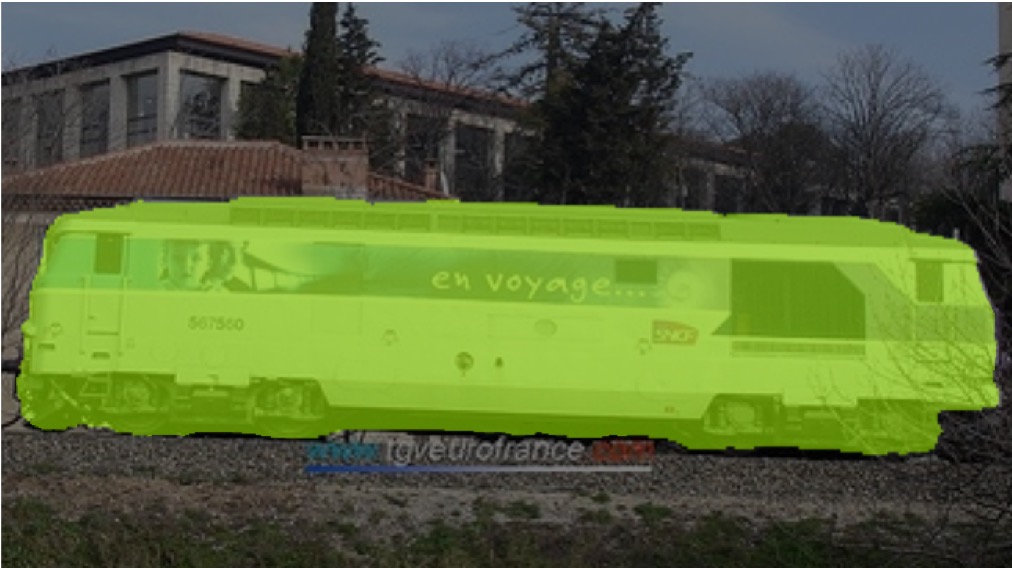}
        \caption{Searched mIoU + BIoU}
    \end{subfigure}
    \vspace{-0.5em}
    \caption{Segmentation results of surrogate losses for mIoU and BIoU.}
    \label{fig-boundary-biou-vis}
\end{figure}

\begin{figure}[tb]
    \centering
    \begin{subfigure}{0.24\textwidth}
        \includegraphics[width=\textwidth]{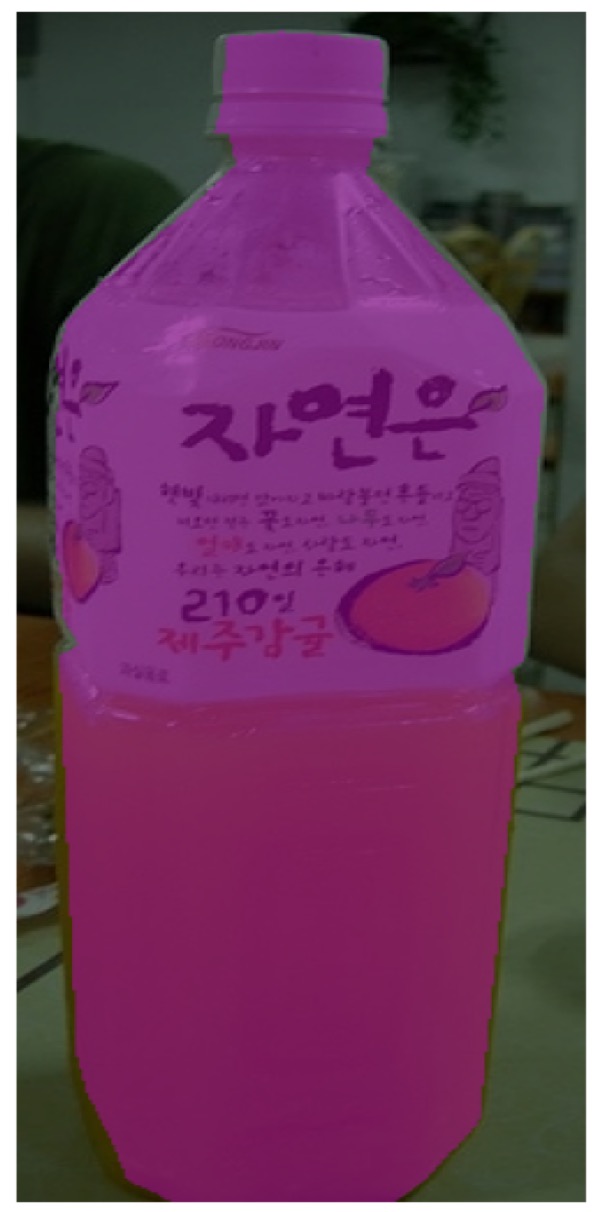}
        \caption{Ground-truth}
    \end{subfigure}
    \begin{subfigure}{0.24\textwidth}
        \includegraphics[width=\textwidth]{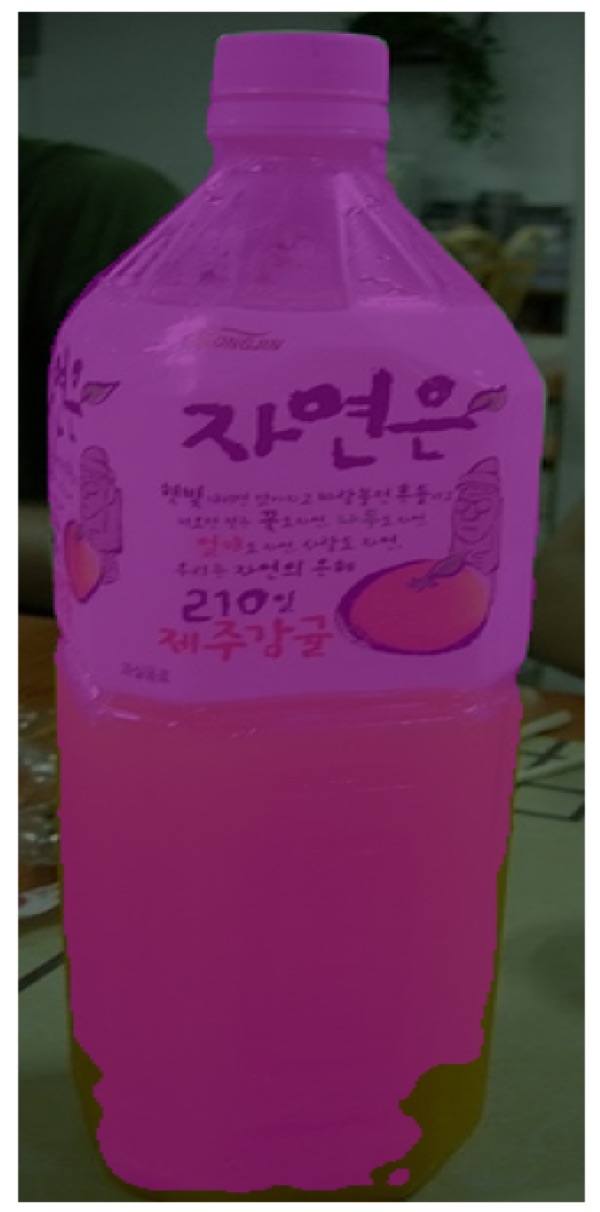}
        \caption{Searched mIoU}
    \end{subfigure}
    \begin{subfigure}{0.24\textwidth}
        \includegraphics[width=\textwidth]{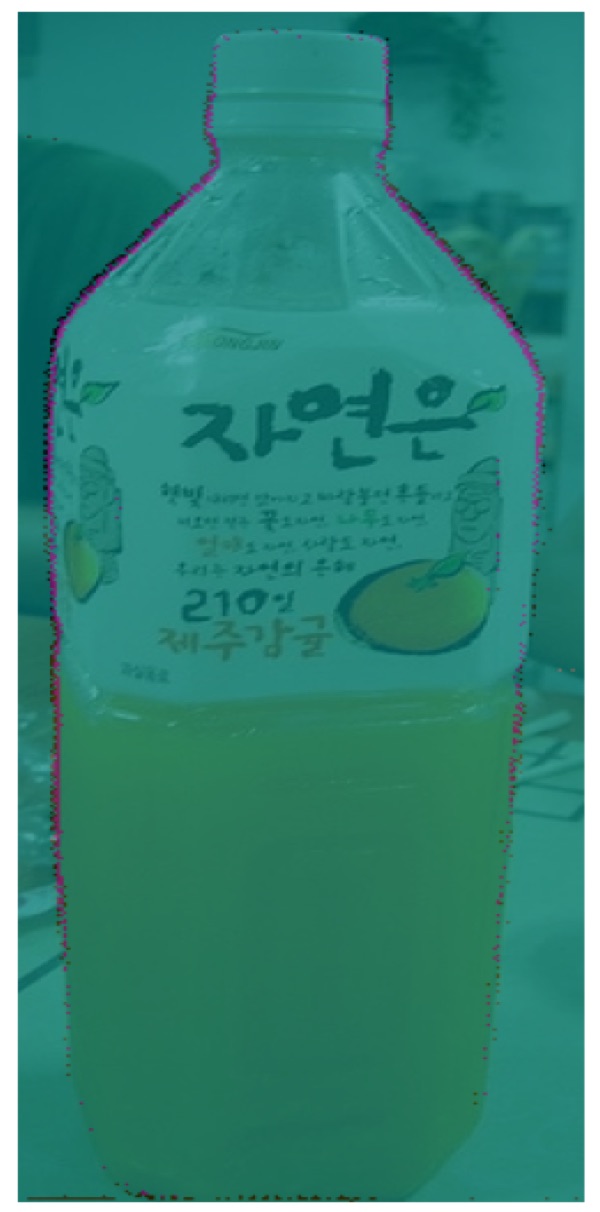}
        \caption{Searched BF1}
    \end{subfigure}
    \begin{subfigure}{0.24\textwidth}
        \includegraphics[width=\textwidth]{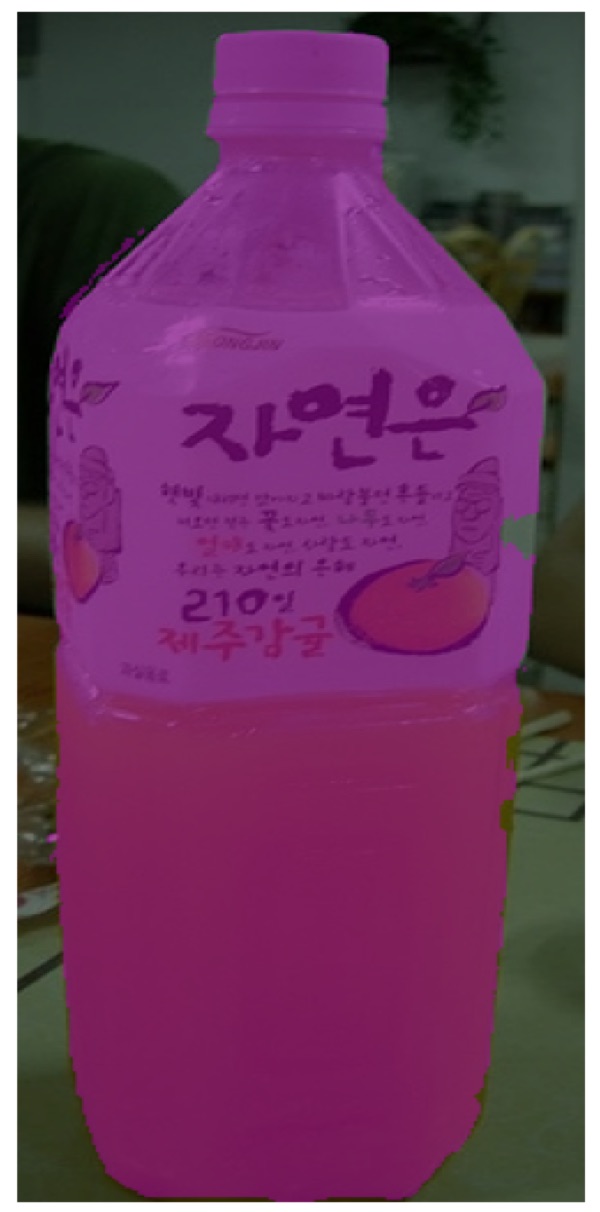}
        \caption{Searched mIoU + BF1}
    \end{subfigure}
    \vspace{-0.5em}
    \caption{Segmentation results of surrogate losses for mIoU and BF1.}
    \vspace{-0.5em}
    \label{fig-boundary-bf1-vis}
\end{figure}

\textbf{Boundary segmentation.} As shown in Table~\ref{exp-voc&city} and Table~\ref{exp-linear-voc&city}, despite the great improvement achieved on BIoU and BF1 scores by training with surrogate losses for BIoU and BF1, respectively, other metrics show a significant drop.
Fig.~\ref{fig-boundary-biou-vis} and Fig.~\ref{fig-boundary-bf1-vis} visualizes the segmentation results of surrogate losses for mIoU, BIoU/BF1, and mIoU + BIoU/BF1.
It can be seen that the surrogate losses for BIoU/BF1 guide the network to focus on object boundaries but ignore other regions, thus fail to meet the needs of other metrics.
Training with surrogate losses for both mIoU and BIoU/BF1 can refine the boundary meanwhile maintain good performance for mIoU.

\begin{figure}[tb]
    \centering
    \begin{subfigure}{0.45\textwidth}
        \includegraphics[width=\textwidth]{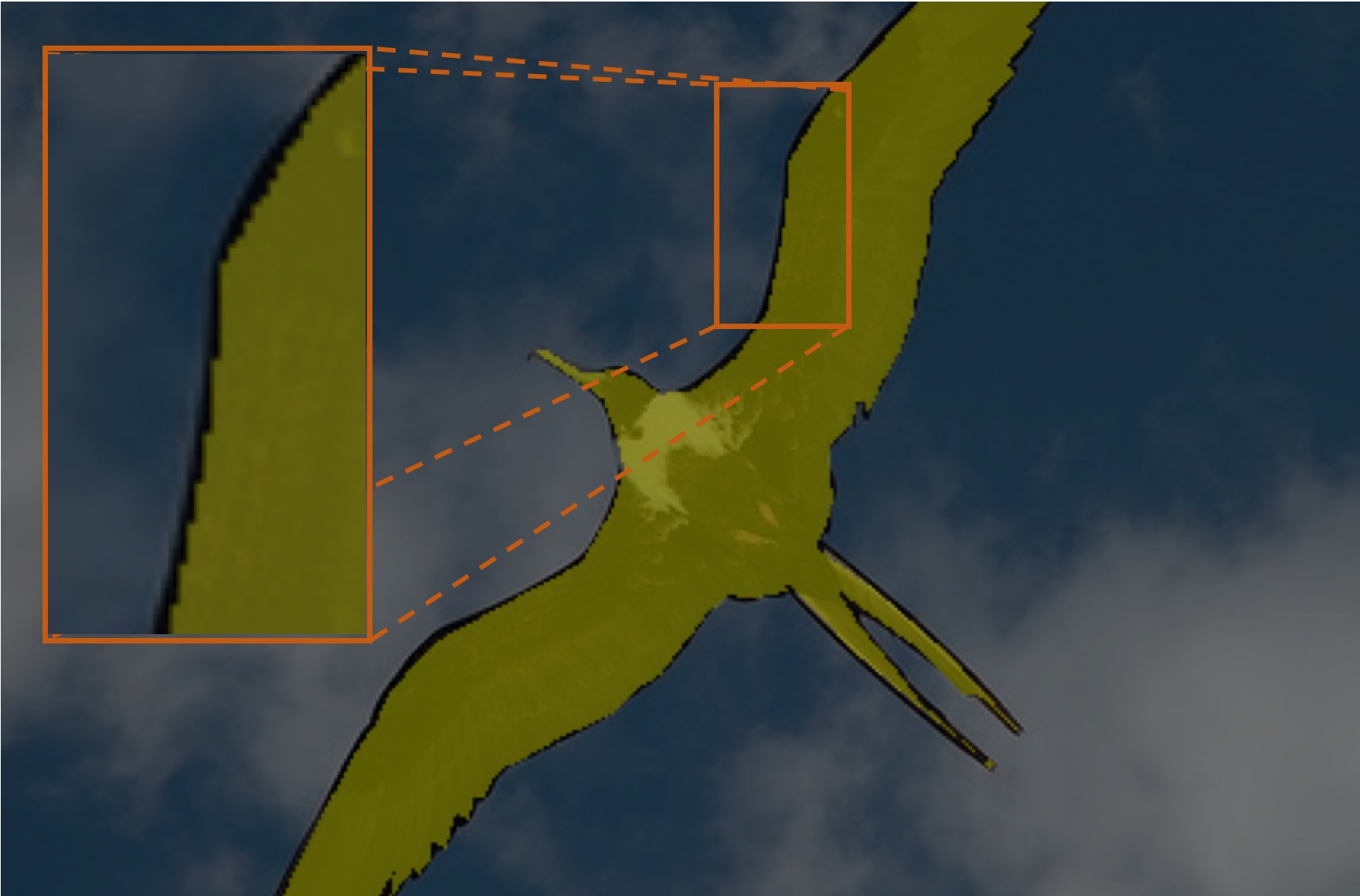}
        \caption{Ground-truth}
    \end{subfigure}
    \begin{subfigure}{0.45\textwidth}
        \includegraphics[width=\textwidth]{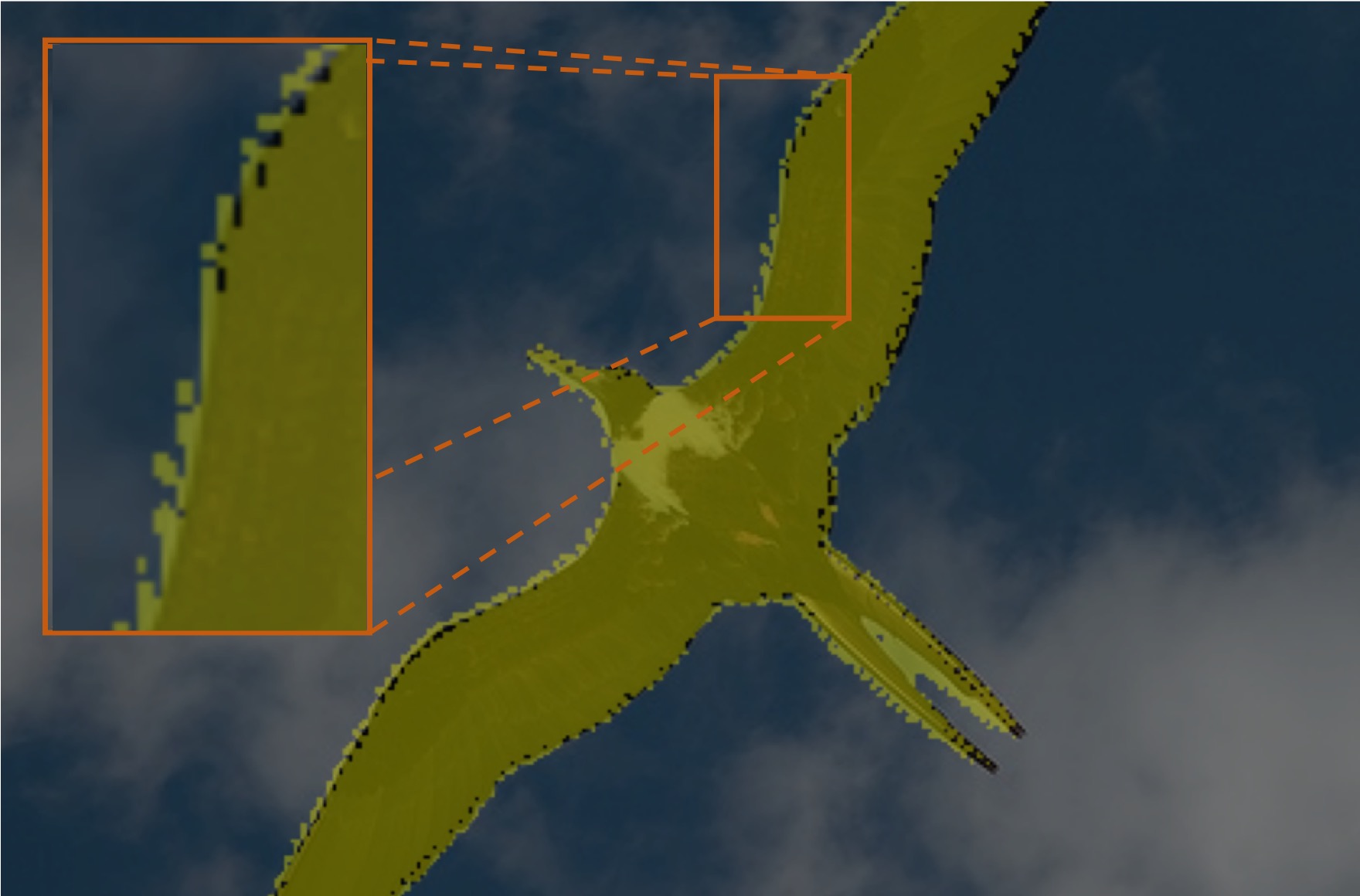}
        \caption{Searched BF1 with tolerance of 5 pixels}
    \end{subfigure}
    \begin{subfigure}{0.45\textwidth}
        \includegraphics[width=\textwidth]{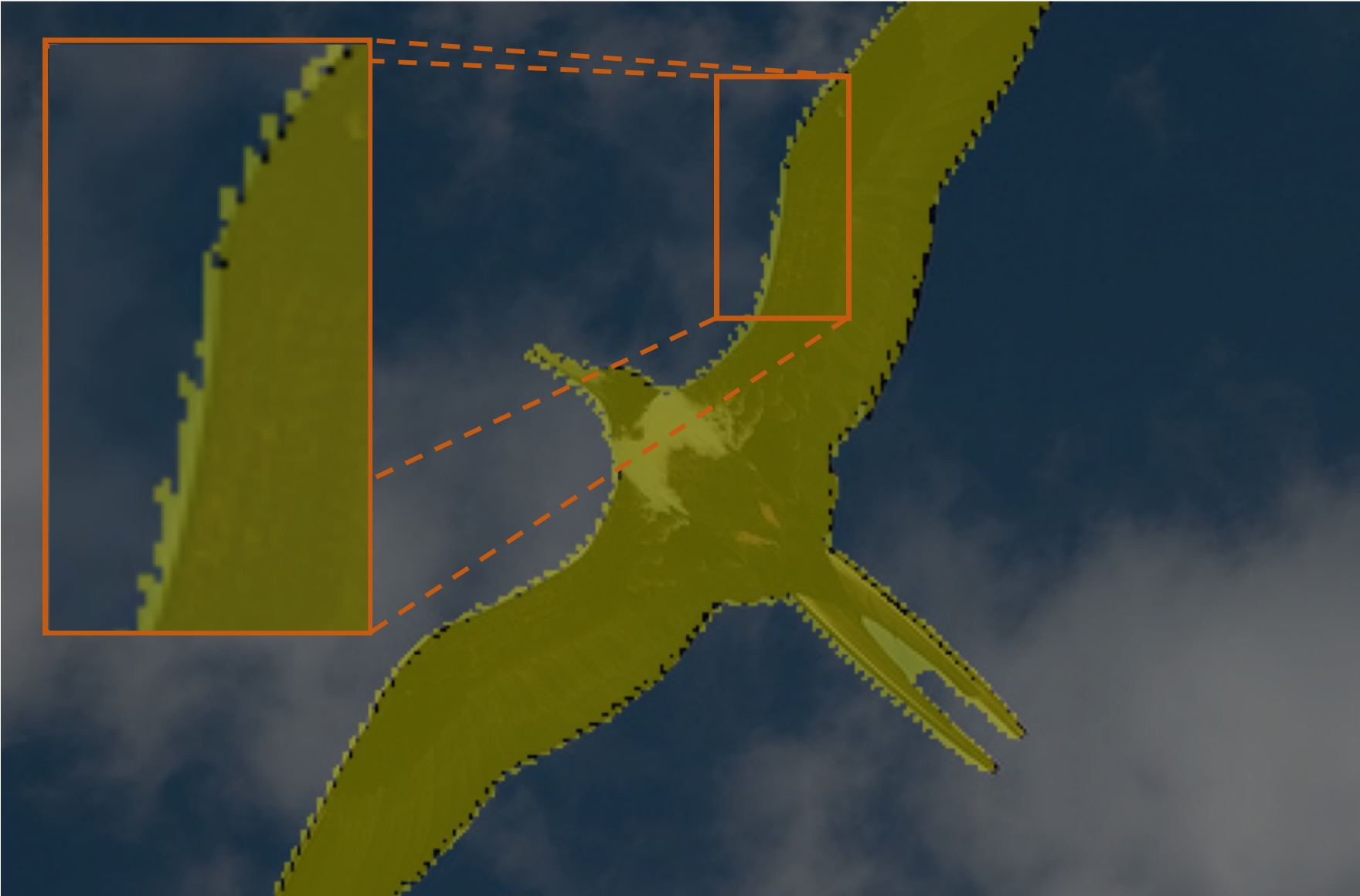}
        \caption{Searched BF1 with tolerance of 2 pixels}
    \end{subfigure}
    \begin{subfigure}{0.45\textwidth}
        \includegraphics[width=\textwidth]{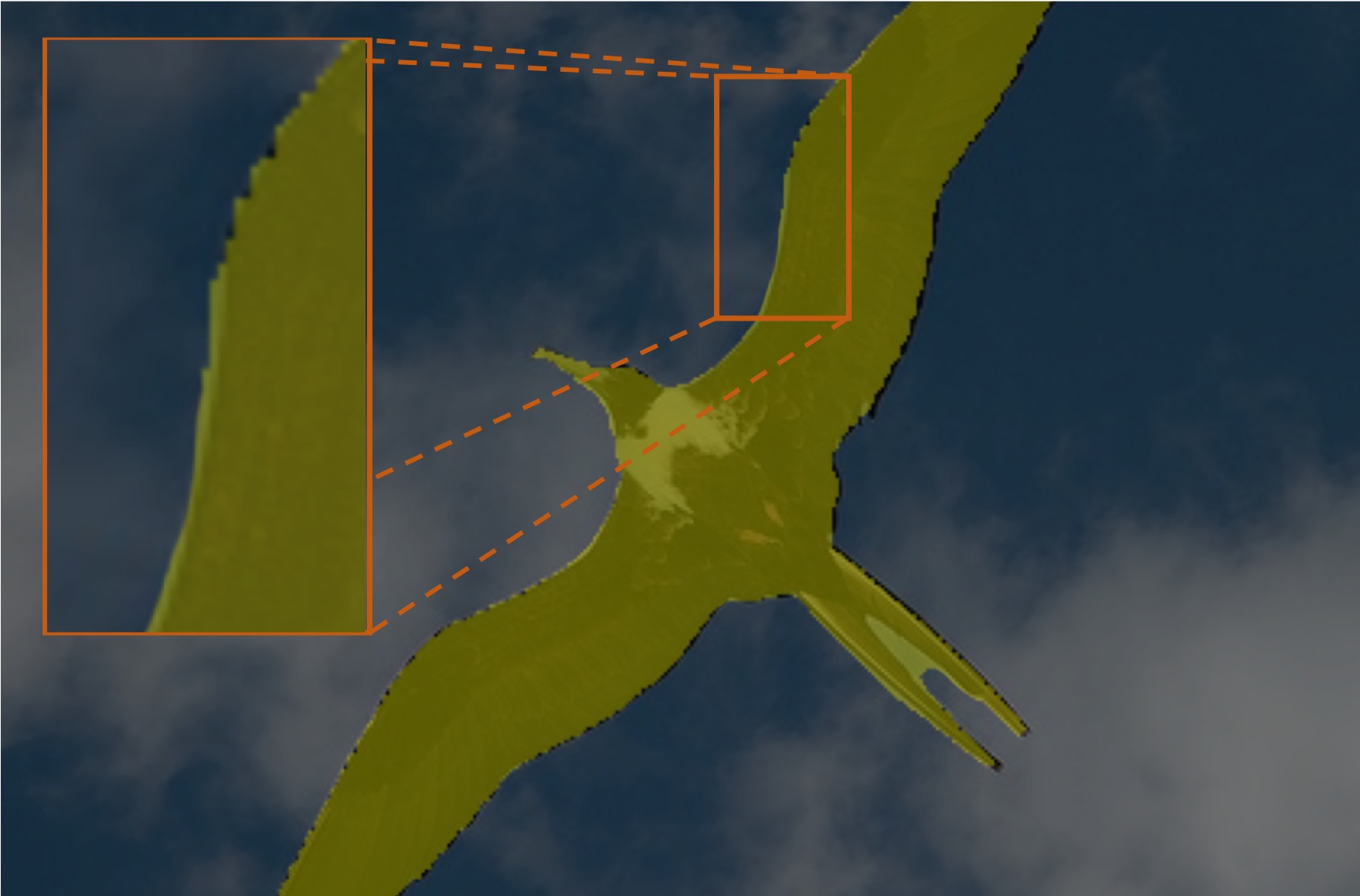}
        \caption{Searched BF1 with tolerance of 0 pixel}
    \end{subfigure}
    \vspace{-0.5em}
    \caption{Segmentation results of surrogate loss for mIoU + BF1, with different BF1 tolerance ranges.}
    \vspace{-0.5em}
    \label{fig-boundary-tol}
\end{figure}

\textbf{Boundary tolerance of the BF1 metric.} Boundary metrics (e.g., BIoU and BF1) introduce the tolerance for boundary regions to allow slight misalignment in boundary prediction. Interestingly, we find that using the surrogate loss for BF1 with non-zero tolerance will lead to sawtooth around the predicted boundaries, as shown in Fig.~\ref{fig-boundary-tol}. Such sawtooth waves are within the tolerance range, which would not hurt the BF1 scores. When the boundary tolerance range in BF1 score reduces, the sawtooth phenomenon gets punished. The corresponding surrogate losses are learned to remove such sawtooth waves.

\end{document}